\documentclass[10pt,journal,compsoc]{IEEEtran}
\ifCLASSOPTIONcompsoc
  \usepackage[nocompress]{cite}
\else
  \usepackage{cite}
\fi
\ifCLASSINFOpdf
\else
\fi
\usepackage{graphicx}
\usepackage{amssymb}

\newcommand{\Fmat}[0]{{{\bf F}}}

\newcommand{\Hmat}[0]{{{\bf H}}}

\newcommand{\Mmat}[0]{{{\bf M}}}

\newcommand{\Xmat}{{\bf X}}
\newcommand{\Ymat}[0]{{{\bf Y}}}
\newcommand{\Zmat}{{\bf Z}}

\newcommand{\hv}[0]{{\boldsymbol{h}}}

\newcommand{\wv}{\boldsymbol{w}}

\newcommand{\ie}{{\em i.e.}}
\newcommand{\eg}{{\em e.g.}}

\usepackage{colortbl}
\definecolor{mygray}{gray}{.88}
\usepackage{amsthm} 

\usepackage{tablefootnote}
\usepackage{booktabs}
\usepackage{hyperref}
\usepackage{amsmath}
\usepackage{threeparttable}
\usepackage{multirow}
\usepackage{multicol}

\begin{document}
%
\title{Motion-aware Dynamic Graph Neural Network for Video Compressive Sensing}
%
%
%
%

\author{%
  Ruiying Lu* \and Ziheng Cheng* \and Bo Chen \and Xin Yuan \\
  National Laboratory of Radar Signal Processing, Collaborative Innovation Center of Information Sensing and Understanding, Xidian University, Xi'an 710071, China.\\
  \texttt{ruiyinglu\_xidian@163.com,zhcheng@stu.xidian.edu.cn,bchen@xidian.edu.cn} \\
  Westlake University\\
    \texttt{xyuan@westlake.edu.cn} \\
  *Equal contribution\\
}

\author{
Ruiying~Lu*,
Ziheng~Cheng*,
Bo~Chen,~\IEEEmembership{Senior Member,~IEEE,}
and~Xin~Yuan,~\IEEEmembership{Senior Member,~IEEE}
\IEEEcompsocitemizethanks{
\IEEEcompsocthanksitem Ruiying~Lu is with the School of Cyber Engineering, Xidian University, Xi'an 710071, China.
\protect\\
E-mail: luruiying@xidian.edu.cn
\IEEEcompsocthanksitem Ziheng Cheng and Bo Chen are with the National Key Laboratory of Radar Signal Processing, Xidian University, Xi'an 710071, China.
\protect\\
E-mail: zhcheng@stu.xidian.edu.cn\protect\\
bchen@mail.xidian.edu.cn
\IEEEcompsocthanksitem Xin Yuan is with Research Center for Industries of the Future (RCIF) and School of Engineering, Westlake University, Hangzhou, Zhejiang 310030, China. 
\protect\\
E-mail: xyuan@westlake.edu.cn
\IEEEcompsocthanksitem Corresponding authors: Bo Chen and Xin Yuan.
\IEEEcompsocthanksitem ~*~: Equal contribution.}}

\IEEEtitleabstractindextext{%
\begin{abstract}
Video snapshot compressive imaging (SCI) utilizes a 2D detector to capture sequential video frames and compress them into a single measurement. Various reconstruction methods have been developed to recover the high-speed video frames from the snapshot measurement. However, most existing reconstruction methods are incapable of efficiently capturing long-range spatial and temporal dependencies, which are critical for video processing. In this paper, we propose a flexible and robust approach based on the graph neural network (GNN) to efficiently model non-local interactions between pixels in space and time regardless of the distance. Specifically, we develop a motion-aware dynamic GNN for better video representation, i.e., represent each node as the aggregation of relative neighbors under the guidance of frame-by-frame motions,
which consists of motion-aware dynamic sampling, cross-scale node sampling, global knowledge integration, and graph aggregation. Extensive results on both simulation and real data demonstrate both the effectiveness and efficiency of the proposed approach, and the visualization illustrates the intrinsic dynamic sampling operations of our proposed model for boosting the video SCI reconstruction results. The code and model will be released.
\end{abstract}

}

\maketitle

\IEEEdisplaynontitleabstractindextext

%
\IEEEpeerreviewmaketitle

\IEEEraisesectionheading{\section{Introduction}\label{sec:introduction}}

 

\IEEEPARstart{I}{nspired} by compressive sensing (CS)~\cite{cs_Candes06,cs_Candes06randomProj,cs_Donoho06}, various computational imaging systems~\cite{Altmann18Science,Yuan2021_SPM} have been developed to achieve computationally-efficient high-dimensional data capture and processing.
In particular, snapshot compressive imaging (SCI) is one important branch of computational imaging techniques with wide applications~\cite{Hitomi11ICCV, Wagadarikar08CASSI, reddy2011p2c2}, which utilizes a 2-dimensional (2D) camera to capture the desired 3-dimensional (3D) data (videos or hyperspectral images) by imposing modulations and then compressing into a single frame (measurement). 
Different from conventional imaging systems employing direct sampling strategies, SCI systems compress the high-dimensional signals along time (\ie, CACTI~\cite{Patrick13OE, Yuan14CVPR}) or spectrum (\ie, CASSI~\cite{Wagadarikar09CASSI}) into the compressed measurements, leading to promising advantages in acquisition efficiency, storage consumption, and low bandwidth footprints. 

\begin{figure*}[!t]
\centering
\includegraphics[width=1\textwidth]{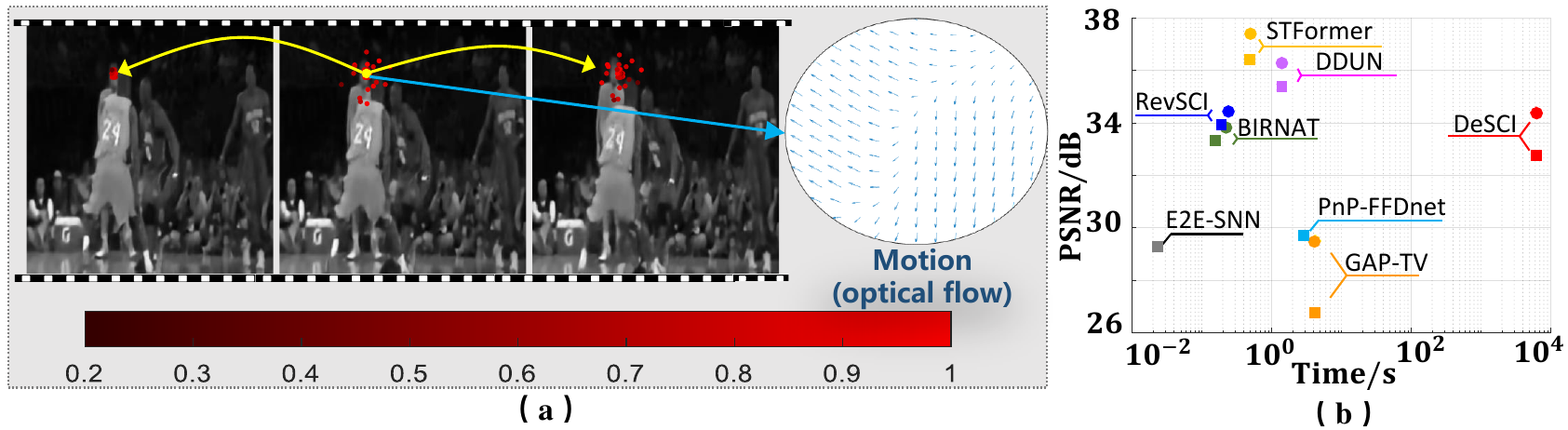}
\caption{\label{fig} (a) An illustration of the spatial-temporal non-local correlations between the central point (yellow point) and related points (red points, dynamically selected by our model) in adjacent three frames, and the motion directions represented by optical flow surrounding the central point. (b) {The comparisons of reconstruction quality and inference time by appending our proposed Graph module to various backbones.} The same color represents the same backbone; '$\circ$'/ '$\qedsymbol$' denote the results with/ without MadyGraph.}
\end{figure*}

Taking the compressed measurements as input, SCI reconstruction algorithms focus on efficiently and effectively recovering the desired video frames.
The first mainstream methods are based on optimization models, which differ in various prior knowledge, \eg, total variation (TV) used in GAP-TV \cite{Yuan16ICIP_GAP} and TwIST \cite{TwIST}, sparsity in GMM-based methods~\cite{Yang14GMM, Yuan15GMM}
, and non-local low-rank in DeSCI \cite{Liu18TPAMI}.
The second mainstream methods are based on deep learning networks, which prefer learning the direct mapping from measurements to video frames, such as off-the-shelf U-net~\cite{Qiao2020_APLP}, 3D convolutional neural network (CNN) based RevSCI~\cite{Cheng2021_CVPR_ReverSCI}, recurrent neural network (RNN) based BIRNAT~\cite{Cheng20ECCV_BIRNAT,cheng2022recurrent}.
Recently, some researchers combine deep networks with optimization methods to reconstruct the desired videos, such as deep unfolding models~\cite{wu2021SCI3D,li2020end, Ma19ICCV,Meng_GAPnet_arxiv2020,meng2023DeepUnfolding,zheng2023unfolding} and plug-and-play algorithms~\cite{Yuan20PnPSCI, PnP_SCI_arxiv2021}.
However, although existing approaches can reconstruct the videos decently, they are limited in efficiently capturing long-range dependencies across video frames, restricting further performance improvement.

{For video SCI reconstruction, capturing non-local interactions is of central importance, since similar pixel patterns frequently recur between distant pixels across space-time~\cite{simonyan2014two,wang2018non}, and the random masks of SCI systems inherently disrupt the local continuity of videos when modulating space-time continuum scenes~\cite{Patrick13OE, Yuan14CVPR}.
However, as the most widely utilized deep learning methods, both the convolutional and recurrent networks build blocks that process one local neighborhood at a time, either in space or time. {While directly employing deep stacks of CNNs or RNNs can naturally extend the receptive field, this strategy is inherently limited in inefficient computation, challenging optimization, and multihop dependency modeling, \ie, difficulty in delivering messages back or forth between distant positions.}
In recent years, the self-attention mechanisms, implemented within the Transformers, CNNs, and RNNs, have been applied to capture the non-local dependencies for SCI~\cite{Cheng20ECCV_BIRNAT, Miao19ICCV, Meng20ECCV_TSAnet,wang2022spatial, Wang_2023_CVPR,wang2023deep}. However, these models are inherently limited in the following perspectives: 1) severe computation overhead due to the $\mathcal{O}\left(N^{2}\right)$ complexity, 2) redundant information by capturing relations across all pairs of pixels, and 3) only performing along the spatial dimension and neglecting the temporal dimension or vice versa. Consequently, there remains a gap in constructing an elaborately designed model to efficiently capture both the local and global spatial-temporal dependencies for better video SCI representation learning.

Bearing these concerns in mind, we provide a graph modeling approach that offers an elegant and principled way to {model correlations} between any two positions, regardless of their spatial-temporal distance.
Intrinsically, instead of relying solely on geometric properties, the graph model enables modeling beyond the constraints of Euclidean spatial distances and captures the connections, interactions, and dependencies among entities. 
With our graph model, {intrinsic correlations in videos can be excavated even if they are not physically obvious in Euclidean space, such as modulation interactions, information flow, and dynamic relations, which might be concealed in traditional distance-based models}. By transcending the constraints of Euclidean space, the graph model is capable of modeling non-local correlations and opens up new possibilities for understanding and analyzing complex scenarios for SCI systems.

Consequently, this paper proposes a motion-aware dynamic graph, dubbed MadyGraph, for video SCI reconstruction. {To alleviate the computational issues while enjoying the advantages brought by graph-structured features, a dynamic graph node sampling scheme is proposed. In contrast to traditional graph modeling with fixed adjacent matrix, we develop a novel data-adaptive frame-wise dynamic sampling to efficiently and accurately select the most relative neighbors for each node.
Furthermore, to better exploit non-local dependency~\cite{wang2018non}, as depicted in Fig.~\ref{fig} (a), we leverage optical flow as guidance into our dynamic sampling strategy.}
To elaborate a bit, we develop MadyGraph by the following concerns: i) capture dependencies regardless of the distance along both the spatial and temporal dimensions, achieved by {\em graph modeling}; ii) flexibly aggregate the most relative contextual information without much redundancy, achieved by {\em dynamic sampling} the neighboring nodes; iii) leverage appropriate priors, such as motion information (by {\em motion-aware dynamic walks}), sparsity and non-local self-similarity (by {\em cross-scale node sampling}), into graph modeling to better exhibit correlations among entities. 

In summary, MadyGraph models non-Euclidean relationships among dynamically selected spatial-temporal related nodes and aggregates non-local information, aiming to enhance feature representation for video reconstruction. The specific contributions can be summarized as follows:
\begin{itemize}

\item 
{We formulate the video SCI reconstruction as a graph learning task and provide an elegant solution by developing a motion-aware dynamic graph network (MadyGraph) to efficiently model non-local spatial-temporal dependencies. To the best of our knowledge, the establishment of a graph model for video SCI reconstruction remains unexplored in this field.}

\item {A novel dynamic sampling strategy is proposed for efficient graph modeling, which alleviates the computational issues by achieving a linear complexity $\mathcal{O}(N)$ in contrast to prevailing non-local modeling approaches with quadratic complexity $\mathcal{O}(N^2)$, aligning with the realistic computational requirements of video SCI.}

\item {We showcase the {effectiveness and efficiency of the proposed MadyGraph for video SCI}, which achieves state-of-the-art (SOTA) performances {concerning} comparison models on both simulation and real datasets.} 

\item {Furthermore, the graph module could be utilized as a flexible and lightweight component, which is capable of “plugging” into any existing SCI reconstruction approaches, leading to considerable improvements as shown in Fig.~\ref{fig} (b).}

\end{itemize}

\section{Video Snapshot Compressive Imaging}
\subsection{Mathematical Model of Video SCI}

In video SCI, the 3D video is modulated by dynamic masks and then compressed into the measurement (a coded 2D frame) along the time dimension, which is decoded later with reconstruction methods to recover the original video.
As depicted in Fig.~\ref{fig: sci model}, $B$ high-speed video frames $\{\Xmat_b\}_{b=1}^B \in {\mathbb R}^{H \times W}$ are modulated by $B$ coding masks $\{\Mmat_b\}_{b=1}^B \in {\mathbb R}^{H \times W}$, and then integrated over time using a camera. Mathematically, the compressed coded measurement frame $\Ymat \in {\mathbb R}^{H \times W}$ can be expressed as
\begin{equation} \label{Eq:YXC}
{\textstyle \Ymat = \sum_{b=1}^B \Xmat_b \odot  \Mmat_b + \Zmat}\,,
\end{equation}
where $\odot$ denotes the Hadamard (element-wise) product and $\Zmat \in {\mathbb R}^{H \times W}$ refers to the system noise.
It has been proved that high-quality reconstruction is achievable when $B>1$~\cite{Jalali19TIT_SCI}.
In this paper, we focus on the inverse problem, \ie, reconstruction of original video frames $\{\Xmat_b\}_{b=1}^B$ given measurement $\Ymat$ and masks $\{\Mmat_b\}_{b=1}^B$.

\begin{figure}
\includegraphics[width=0.485\textwidth]{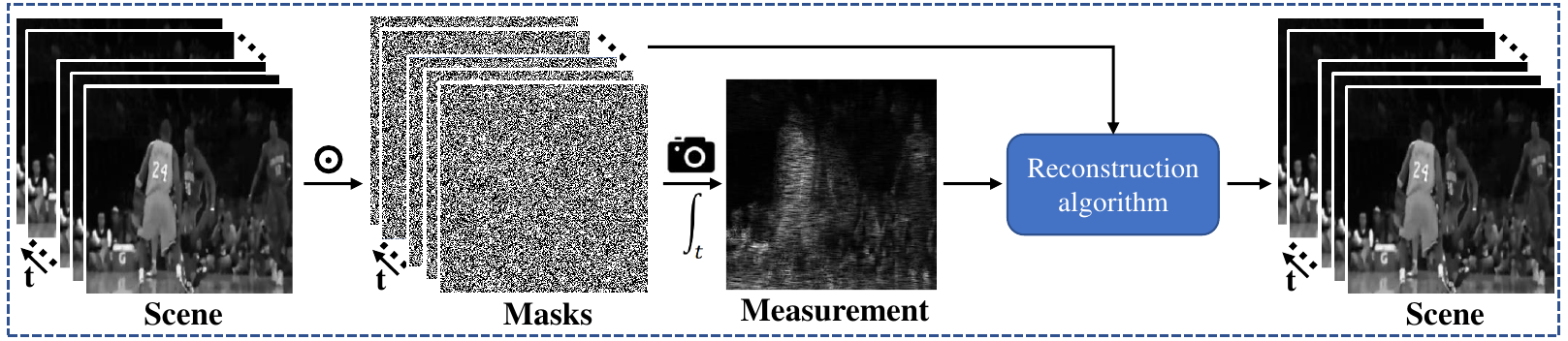}
\caption{\label{fig: sci model} Principle of video SCI system. The original video frames are modulated by dynamic masks and then compressed by the camera into a snapshot measurement, then decoded by reconstruction algorithms to recover the video.}
\end{figure}

\subsection{Related Work}
\noindent{\textbf{Video Snapshot Compressive Imaging:}} Video SCI is a hardware-encoder-plus-software-decoder system~\cite{Yuan2021_SPM}. 
In the encoder, the critical component is the spatial light modulators for effectively modulating the original scenes, usually using a physical mask~\cite{Patrick13OE, Yuan14CVPR}, or different patterns on the digital mirror device (DMD)~\cite{Hitomi11ICCV, Qiao2020_APLP, Ma2021_LeSTI_OE, Qiao2020_CACTI, Qiao2021_MicroCACTI, Sun17OE}.
Regarding the decoder, some optimization-based methods introduce different priors to recover the desired videos by iteratively optimizing, \eg, GAP-TV~\cite{Yuan16ICIP_GAP}, GMM~\cite{Yang14GMM, Yuan15GMM}, and DeSCI~\cite{Liu18TPAMI}, however suffering from the extremely long and impractical reconstruction time. In recent years, a boost in SCI reconstruction efficiency was achieved by introducing deep learning-based methods~\cite{Cheng2021_CVPR_ReverSCI, Cheng20ECCV_BIRNAT,wang2021metasci, Meng_GAPnet_arxiv2020,li2020end, Wang_2023_CVPR}, which can recover the high-quality videos within tens or hundreds of milliseconds.
Most recently, a dense deep unfolding network (DDUN)~\cite{wu2021SCI3D} is designed with 3D CNN prior, which combines the merits of both optimization-based and deep learning-based methods, achieving impressive performance in video SCI.
Nevertheless, these approaches usually neglect the long-range dependencies over the space-time volume during reconstruction, leading to a potential requirement for efficient non-local modeling.
Recently, STFormer~\cite{wang2022spatial} is proposed to exploit the correlation in both spatial and temporal domains, composed of a spatially local self-attention branch, a temporal self-attention branch, and a fusion network, which achieves competitive performance. However, STFormer cannot simultaneously capture the spatial-temporal correlation and suffers from expensive computation costs with redundant correlation modeling.

\noindent{\textbf{GNN for Non-local Image/Video Processing:}} For a natural image/video, similar pixel patterns frequently recur so that many non-local methods have shown pleasant performance in different tasks~\cite{Buades05NLM, Dabov07BM3D}.
The non-local operation computes the response at a position as a weighted sum of the features at all positions, across
space, time, or space-time~\cite{wang2018non,liu2018non,zhang2019residual}, but suffering from the exhausting computation and information redundancy.
Graph neural network (GNN) based methods, which
propagate information along graph-structured input data, can capture non-Euclidean non-local dependencies and simultaneously alleviate the computational issues.
{Researchers also have explored GNNs in video understanding tasks, including video action recognition~\cite{wang2018videos},
temporal action localization~\cite{xu2020g}, trajectory prediction~\cite{an2022dginet}, and video object
segmentatio~\cite{lu2020video}.
For example, some recent works~\cite{gkalelis2021objectgraphs} adopt graph reasoning (GR) for video content understanding, but mostly limit GR to a single frame. MAGRH\cite{zeng2022motion} design a dual-branch architecture to extract the frame-wise and clip-wise hash codes for video retrieval, however, still limiting the motion information within the clip. In a word, prevailing methodologies typically focus on either capturing partial dependencies at the frame level or local contexts across consecutive frames at the regional level~\cite{chen2022survey}.
Effectively and efficiently capturing long-range global dependencies or crucial contexts without redundant information remains further exploration.}
The most related works of our proposed model are GraphSAGE~\cite{HamiltonYL17} and DGMN~\cite{Zhang0AT20}, which utilize sampled graph nodes to capture position-based context. However, GraphSAGE~\cite{HamiltonYL17} simply samples nodes with fixed positions along the spatial dimension, independent of the actual input. Absorbing the ideas from deformable convolution~\cite{dai2017deformable}, DGMN~\cite{Zhang0AT20} adaptively samples graph nodes for message passing according to the input. Nevertheless, it is limited to the 2D spatial dimension without exploring the temporal dimension. Furthermore, in DGMN, the position of each node is estimated only based on the features of a set of sampled nodes without checking the whole feature map and explicit guidance, resulting in insufficient information exploration. Different from it, we extend the non-local modeling to space and time while keeping a lightweight computational cost, and dynamically sample related nodes not only according to the whole feature map but also taking the knowledge of motion information between frames into consideration. {Moreover, as the physical mask of the SCI system leads to the changing light transmittance of different positions across different frames, we make particular designations such as input re-masking and frame-wise sampling to fit the SCI problem.}

\section{The Proposed Model}

\begin{figure*}[t!]
\includegraphics[width=1.0\textwidth]{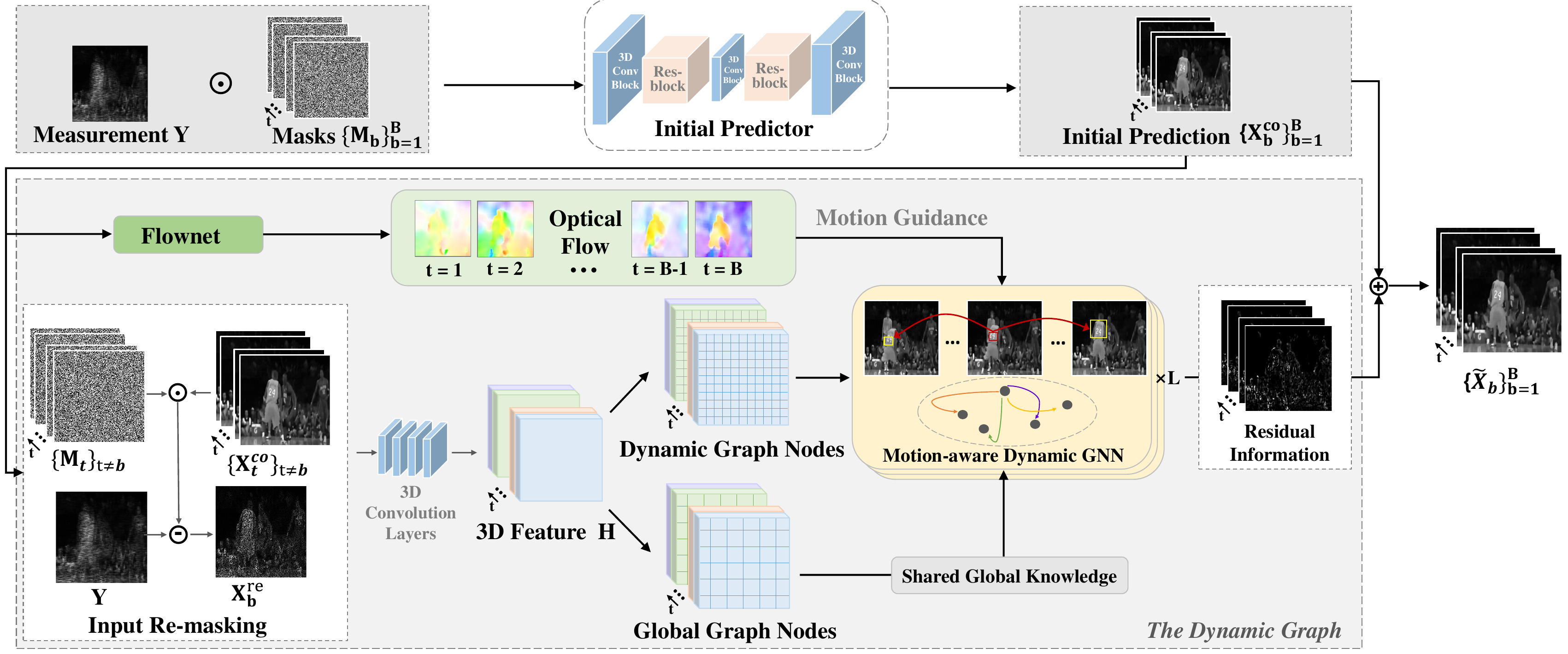}
\caption{\label{fig:model} {Illustration of the whole framework of MadyGraph. The top row of the network takes measurements and masks as input and outputs the initial candidate video through the initial predictor, composed of three 3D convolution blocks and two residual blocks. The bottom network refers to the dynamic graph module, which starts by re-masking and mapping the initial prediction through a 4-layer 3D convolution network, then updates the video representation through graph learning, involving motion-aware dynamic graph sampling and global knowledge integration, and finally aggregates features and outputs the residual details. To this end, the initial prediction and residual information are combined together to achieve the high-quality video.}}
\end{figure*}

Given the compressed measurement $\Ymat$ captured by the SCI system and the corresponding masks $\{\Mmat_b\}_{b=1}^B$, our goal is to output a fine-grained reconstructed video. 
One of the main drawbacks of the traditional deep learning-based methods is the computational cost of training from scratch, which may require up to tens of thousands of iterations and weeks of time for training. As such, this work introduces a new training strategy accomplished with a residual framework and aims to complement the details of the initial prediction, which is often possible to be cheaply obtained. Specifically, in our proposed framework as depicted in Fig.~\ref{fig:model}, we incorporate an initial predictor (top row of Fig.~\ref{fig:model}) to provide a data-adaptive candidate for the fine-grained video. Subsequently, the dynamic graph network (bottom row of Fig.~\ref{fig:model}) focuses on modeling the residual part.


Overall, our proposed MadyGraph illustrated in Fig.~\ref{fig:model}, consists of four steps: $i$) The predicted candidate video frames are obtained from the initial predictor, and then re-masked and transmitted to the 3-D features. $ii$) The dynamic graph is constructed to model the non-local space-time correlations in videos with two mechanisms: motion-aware dynamic sampling and cross-scale node sampling. $iii$) The dynamically constructed graph is aggregated in the feature domain and then mapped to the desired residual video details. $iv$) The residual details are added to the initial prediction to reconstruct the final fine-grained video.

\subsection{Initial Prediction}
We introduce a simple residual framework to reduce the training cost and difficulty, which is inspired by the principles of predictive coding~\cite{rao1999predictive} and neural compression algorithms~\cite{yang2021insights}, suggesting that residual frameworks are easier to model than one-shot reconstruction~\cite{yang2022diffusion, marino2020improving}. Specifically, we first utilize an initial predictor to provide a data-adaptive predicted candidate video. Then the dynamic graph network only needs to model the residual details as replenishment. 
{For fast training, we construct a lightweight initial predictor to give the initial prediction, composed of three 3-D convolution blocks and two residual blocks, taking the frame-wise dot product of measurement and masks as the input.}

{The initial predictor network provides a data-adaptive candidate video for the residual network, \ie, dynamic graph module, to add further complement information, thus significantly reducing the burden of the graph network.} 
Note that we optimize the dynamic graph module with this simple initial predictor only at the training stage, and it is flexible to directly append the well-trained dynamic graph module to other backbone models at the testing stage.} In our paper, the backbone models could be any existing video SCI models including both the optimization-based and deep learning-based models. Surprisingly but reasonably, the reconstructed videos of different backbone models can be significantly improved even without time-consuming fine-tuning. 
{Next, we will mainly focus on the dynamic graph module to learn the residual information.}

\subsection{Mask-aware Feature Extraction}

After obtaining the initial predicted video, we designed a dynamic graph network to learn the complementary detailed information. 
{In terms of the network input for predicting the complementary information, one straightforward way is to directly use the reconstructed videos from the initial predictor as inputs. 
However, after extensive experiments, we found it is difficult to obtain good results in this manner, since critical information always be smoothed out in the initial reconstructed videos. Furthermore, directly taking the initial predicted video as input gains no more information increment to learn the complementary details. Recalling that the physical mask is used to modulate the video frames in the snapshot compressive imaging system, thus, the light transmittance of different positions on the mask varies across different frames of videos. As a result, the fidelity of the modulated video information is position-sensitive.
This observation inspires us that the modulation mask could be used as guidance to direct the model to focus on regions with high-fidelity representations.}
In order to explore the fine-grained details, we should add more primordial information about SCI systems such as masks and measurements.
Therefore, we develop a {\em re-masking operation} to better fit video SCI and make good use of the critical knowledge of hardware systems for SCI decoding.

Specifically, based on the initial predicted video $\{\Xmat_b^{Init}\}_{b=1}^B \in {\mathbb R}^{H \times W}$, we re-mask each frame to enrich information with the measurement $\Ymat$ and mask-aware re-modulated frames ${\Xmat_t^{Init}} \odot {\Mmat_t}$. This step aims to incorporate the complementary information that might be neglected or smoothed in the initial predictor.
Recalling the mathematical model of video SCI in Eq.~\eqref{Eq:YXC}, we re-mask each frame by:
\begin{equation} \label{Eq:remask}
\textstyle{\Xmat_b^{re} =  \Ymat -\sum_{t=1, t \neq b}^{B} ({\Xmat_t^{Init}} \odot {\Mmat_t}),}
\end{equation}
where $\Xmat_b^{re}\in {\mathbb R}^{H \times W}$ denotes the $b^{th}$ re-masked frame, acquired by {subtracting} the summation of other masked frames from the measurement. Thus, in the re-masking operation of each frame, all the inputs of the original system and the initial video are incorporated to enrich its information, facilitating further feature extraction. 
Furthermore, the re-masking operation encourages the model to better capture dependencies between the masked position and unmasked context positions by recovering the masked features from the contextual information.

With the re-masked frames $\{\Xmat_b^{re}\}_{b=1}^B$ as input, a 4-layer 3D convolution network $\mathcal{F}_e$ is constructed to encode the data cube into high-dimensional features, as:
\begin{equation} \label{Eq:input_feature}
{\textstyle \Hmat = \mathcal{F}_e(\{\Xmat_b^{re}\}_{b=1}^B)},
\end{equation}
where $\Hmat \in {\mathbb R}^{H\times W \times B \times C}$ is a 4D tensor, $C$ is the channel dimension, and $B$ refers to the temporal dimension (the number of video frames). After the 3D convolution network, the encoded feature of each frame is a 3D tensor, {in which} the representation of each pixel at a single frame corresponds to a vector with the dimension of $ {\mathbb R}^C$.

\begin{figure*}[t!]
\includegraphics[width=0.9\textwidth]{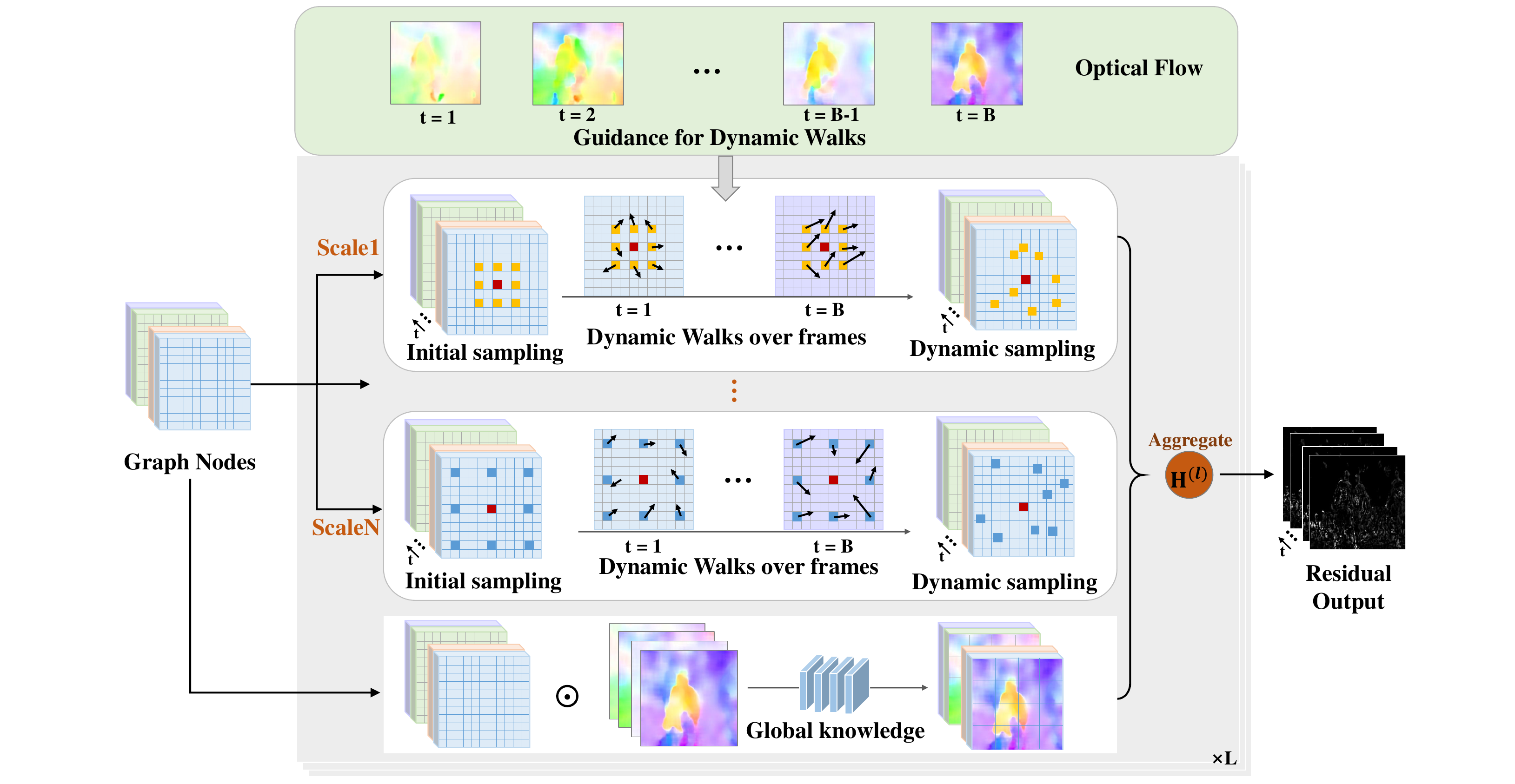}
\centering
\caption{\label{fig:gnnmodel} {Illustration of the sampling strategy and aggregation in our dynamic graph module}: 1) The rows from scale 1 to scale N refer to N-cross-scale sampling pipelines. 2) In each pipeline, firstly, the initial sampling is uniformly distributed; then, the dynamic walks of B frames are generated guided by optical flow; finally, the dynamic walks are performed upon the initially sampled positions for dynamic sampling. 3) The bottom row refers to the global semantic-level graph nodes. 4) To this end, the dynamic graph nodes are aggregated and mapped to form the residual fine-grained details of videos.}
\end{figure*}

\subsection{Motion-aware Dynamic Graph Modeling}

\subsubsection{Graph Definition and Notation}
Given the feature maps $\Hmat \in {\mathbb R}^{H\times W \times B \times C}$, where $B$ is the number of frames and $C$ is the feature dimension, our goal is to update the video feature map {with intrinsic} structured information in the graph domain. 
In order to learn the structured correlations, we construct a feature graph $\mathcal{G} = \{\mathcal{V}, \mathcal{\epsilon}, \mathcal{A}\}$, where $\mathcal{V}$ denotes a set of node features, $\mathcal{\epsilon}$ {refers to} the edges, and $\mathcal{A} \in \mathbb{R}^{|\mathcal{V}|\times|\mathcal{V}|}$ is the adjacency matrix, representing the connection relations between nodes. 
In this work, we first construct the undirected graph and then update the graph representations with the adaptively learnable adjacency matrix for downstream video SCI in each self-loop. {In the following, we will start with the definition and construction of the graph for video SCI, then specifically introduce our proposed dynamic node sampling strategy.} 

\textbf{Nodes of Graph:}
One straightforward way is to {recognize} the feature vectors of each pixel as the nodes of the graph, however, which may cause the following problems: 1) The computational cost is significant due to the large number of nodes being $H\times W \times B$. The corresponding computation complexity is quadratic to the node number as $O(H\times W \times B)^2$, which occupies unaffordable time and memory resources {when} handling videos with large spatial resolutions or long duration; 2) If we compute the relations across all pairs of pixel nodes, the captured information tend to be redundant; 3) If we only sparsely select several nodes to be related to the current node, the available pixel-level information becomes limited, making it challenging to capture and express the structured and semantic relationships present in the video.
To solve the above problems, we define initial graph nodes of our video graph as patches of size $p \times p$ in every frame in the 3D latent feature map $\Hmat$, and develop a dynamically sparse and multi-scale sampling strategy to abandon the redundant information. This {leads to} a total number of graph nodes as $|\mathcal{V}| = {\frac{H}{p} \times \frac{W}{p} \times B}$.
In this way, the computational consumption problem can be relieved and the structured and semantic relations across space and time in videos can be well excavated. 

\textbf{Graph message passing:} The graph representation can be updated through the aggregation operation. The graph message passing phase usually takes $L$ iteration steps to update node representations. Namely, the representation of each node only depends on its $L$-hop neighborhoods.
Formally, at each iteration, the graph representation aggregation can be described as a weighted summation, as:
\begin{equation} \label{Eq:graph}
{\textstyle  \hv_i^{(l+1)} = \sum_{j \in \mathcal{N}(i)} \mathcal{A}_{i,j} \mathcal{T}(\hv_j^{(l)})},
\end{equation}
where $\hv_i^{(l+1)}$ represents the $i^{th}$ node feature at the $(l+1)^{th}$ iteration, computed by weighted summation of the node features $\hv_j^{(l)}$ at the previous iteration. The $\mathcal{A}_{i,j}$ represents the connection relationship between nodes $\hv_i^{(l)}$ and $\hv_j^{(l)}$, and $\mathcal{N}(i)$ refers to the self-included neighbour of the node $\hv_i^{(l)}$. Here, $\mathcal{T}$ is a transmit operation imposed on $\hv_i^{(l)}$.
In this work, the initial node features $\hv_i^{(0)}$ are {obtained} from the initial input feature maps as $\Hmat^{(0)}=Conv^s(\Hmat) \in {\mathbb R}^{\frac{H}{p}\times \frac{W}{p} \times B \times C}$, where $Conv^s$ denotes a 2D $p \times p$ strided convolution operation. This indicates the initial node features of each $p \times p$ patch are represented as a vector $\hv_i^{(0)} \in {\mathbb R}^{C}$.

{Given the set of nodes, $\mathcal{V}$, the subsequent important question is how to efficiently construct adjacent matrix $\mathcal{A}$ which determines the correlations between graph nodes.} When analyzing videos, both the spatial and temporal relations should be taken into consideration regardless of the limitations of their Euclidean distance. A direct implementation is constructing a fully-connected graph by computing non-local interactions $\mathcal{A}_{i,j}$ between every two nodes across space and time, \ie, $\mathcal{N}(i) = \mathcal{V}$. However, a fully-connected graph often suffers from prohibitively expensive computation, redundant information, and optimization difficulty especially when dealing with limited training data.
Therefore, we construct a {\em dynamic sampling} scheme to dynamically sample a small subset of the time-spacial relevant feature nodes as $\mathcal{N}(i)$, including {\em motion-aware 
dynamic sampling} and {\em cross-scale node sampling}. 

In this way, the node correlations in $\mathcal{A}$ are non-zero only at the positions of selected relevant nodes, imposed with the sparsity prior. This allows for the efficient gathering of non-local dependencies without excessive computations and redundant information. In summary, we aim to design a position-adaptive, weight-learnable, motion-aware, and computation-economy construction of $\mathcal{A}$ for the video graph.

\subsubsection{Motion-aware Dynamic Sampling}
In this part, we introduce a motion-aware dynamic node sampling scheme to adaptively select neighbors $\mathcal{N}(i)$ of each node, {composed of three steps: 1) the initial sampling, 2) the frame-by-frame motion extraction, and 3) the motion-aware dynamic walks.}

\noindent{\textbf{Initial sampling:}} As shown in Fig.~\ref{fig:gnnmodel}, {along} the spatial dimension, we uniformly sample $K$ neighboring nodes from $\mathcal{V}$ for each graph node $\hv_i^{(l)}$, which is a commonly used strategy for graph node sampling~\cite{Zhang0AT20,leskovec2006sampling} based on Monte-Carlo estimation. Along the temporal dimension, we continuously sample neighboring nodes with the same spatial position.
Thus, the initial neighbors $\mathcal{N}(i)$ of each node contain $K \times B$ elements.
The initial sampling scheme, being a position-fixed locally-connected mechanism, neglects the original feature distribution and may overlook important contextual information.
Thus, taking the various data distributions into consideration, we develop a{\em position-specific dynamic walk} mechanism upon the uniform sampling to find the data-adaptive neighbors for each node, which is guided by the motions in videos.

\noindent{\textbf{Motion extraction:}} In order to explicitly learn the dynamic motions across video frames, we use optical flow to represent the frame-by-frame motions to help each node find its most related neighboring nodes. Considering the efficiency and accuracy, a pre-trained lightweight network, $Flownet$~\cite{hui2018liteflownet}, is employed as our optical flow extractor.
Taking the $b^{th}$ frame and the $(b+1)^{th}$ frame of the initial predicted video as inputs, the motion can be extracted by:
\begin{equation} \label{Eq:flow}
{\Fmat_b = Flownet(\Xmat^{Init}_b,\Xmat^{Init}_{b+1}), where \; b=1,..., B-1,}
\end{equation}
where $\Fmat_b$ denotes the optical flow of the $b^{th}$ pair of adjacent frames output from $Flownet$~\cite{hui2018liteflownet}, which consists of a vertical and a horizontal component. Furthermore, we define the motion of the last frame in a video as $\Fmat_B = Flownet(\Xmat^{Init}_B,\Xmat^{Init}_{B-1})$, to provide the motion information in the reverse direction. In the following, we will introduce how to utilize the motion information to guide dynamic sampling, which learns the dynamic walks around initial sampling nodes for each graph node.

\noindent{\textbf{Motion-Aware Dynamic Walks:}} 
To take all the node features and motions into account during dynamic sampling, we absorb the ideas from deformable CNN~\cite{dai2017deformable} and propose a motion-aware dynamic walk upon the initial sampled nodes. As illustrated in Fig.~\ref{fig:gnnmodel}, the black arrows refer to the dynamic walks around initial sampling nodes, which are predicted in a data-driven fashion being aware of the overall feature maps and motion distributions. {Considering the positions of related nodes keep changing over time in a video due to the moving objects and different light transmittance at various positions across frames, we perform the dynamic walks independently for each frame along the spatial dimension. In other words, the dynamic walks vary and adapt with respect to different frames.} 

Considering that dynamic walks are conducted across the 2D spatial domain in each frame, let $\triangle \mathbf{p}_{j,b} \in {\mathbb R}^{2 \times 1}$ denote the predicted walk according to one of the initially sampled neighboring nodes $\hv_j \in \mathcal{N}(i)$ in the $b^{th}$ video frame.
The node walks are predicted by applying a stride convolution layer over the input feature maps and motion representations, \ie,
\begin{equation} \label{Eq:walk}
\triangle \mathbf{P}_{b} = Conv([\Hmat,\Fmat_{b-1},\Fmat_b,\Fmat_{b+1}]_c),
\end{equation}
where $\triangle \mathbf{P}_{b} \in {\mathbb R}^{\frac{H}{p} \times \frac{W}{p} \times (K \times 2)}$ refers to the horizontal and vertical walks (offset) of the $K$ neighboring nodes of all the $\frac{H}{p} \times \frac{W}{p}$ nodes in the $b^{th}$ frame, and $[~]_c$ refers to the concatenation operation along channels. Hereafter, we omit the superscript $(l)$ of iterations for brevity. Intuitively, it will be much easier for the neural network to predict the position of related nodes if the explicit motion of the current node is known. Thus we utilize the optical flow of adjacent frames $\{\Fmat_{b-1}, \Fmat_{b}, \Fmat_{b+1}\}$ to provide the motion information of both the former and latter frames as guidance to better learn the dynamic walks. {Finally, the predicted dynamic walk $\triangle \mathbf{p}_{j,b}$ is executed on the initially sampled position $\mathbf{p}_{j_0,b}$ through the addition of $\mathbf{p}_{j_0,b}$ to $\triangle \mathbf{p}_{j,b}$, both horizontally and vertically.} Then the features are calculated by:
\begin{equation} \label{Eq:walk2}
\hv_{j,b} = \Upsilon(\mathbf{p}_{j_0,b}, \triangle \mathbf{p}_{j,b}, \Hmat_b),
\end{equation}
where $\Upsilon$ denotes the bi-linear sampler~\cite{dai2017deformable} imposed on the original feature map $\Hmat_b$, to obtain the bi-linear interpolated features for dynamic sampling nodes. This is due to the dynamic walks are typically fractional, always resulting in irregular sampling positions; please refer to~\cite{dai2017deformable} for more details about the bi-linear sampler.
In short, by performing motion-aware dynamic walks upon the initially sampled neighboring nodes, we can enhance the process of identifying the most relevant neighbors and extracting more informative features for the graph nodes.

\subsubsection{Cross-Scale Node Sampling}

{To capture longer dependencies for each node effectively, we employ a cross-scale node sampling mechanism to expand the receptive field size, while maintaining computational efficiency through dynamic sampling.} Specifically, different dilation rates (denoted as $d$) are used to sample neighboring nodes of various distances whilst maintaining a small number of connected nodes.
As shown in Fig.~\ref{fig:gnnmodel}, composing different dilation rates can find related nodes across various scales, which could be utilized in conjunction with the dynamic walks. 
{A limitation of this parallel architecture is that the computation memory will grow in proportion to the increasing scales, however, which is cost-effective due to its sparsity and substantial performance gains it delivers.}

\subsubsection{Global Knowledge Integration}

The dynamic sampling strategy mainly focuses on the low-level texture characteristics, which might ignore some high-level semantic information. Therefore, we design a mechanism to extract the semantic-level information and integrate the global knowledge into graph learning. Considering the optical flow can well distinguish the moving objects and the motionless background, it is reasonable to embed optical flow into the feature maps for the focus of foreground objects, as:
\begin{equation} \label{Eq:global}
\hv^{global} = Conv(\Fmat \odot {\Hmat}) ,
\end{equation}
where $\Fmat=[\Fmat_1,...,\Fmat_B] \in {\mathbb R}^{H\times W \times B}$ {re-weights} $\Hmat \in {\mathbb R}^{H\times W \times B \times C}$ by position-wise dot production. Then a 3D stride convolution block with a big step size is employed to extract global knowledge, resulting in a small number of global nodes.
All global nodes are considered as neighboring nodes, contributing to the learning process of each node’s features. 

By leveraging higher-level abstraction, graph nodes are able to not only identify and connect with relevant low-level nodes but also gain a holistic semantic understanding of the entire video at a higher level. Furthermore, the extra computational cost remains insignificant due to: 1) the optical flow has already been computed, and 2) the increase in computational cost due to the additional global nodes is minimal in comparison to the existing low-level nodes. In summary, the global nodes correspond to  high-level visual concepts and could provide complementary information for better graph representation learning.

\subsection{Graph Aggregation and Video Reconstruction}
So far, the adjacent matrix of the graph, which represents the connections between the cross-scale motion-aware dynamic sampled nodes, has been successfully constructed. 
Subsequently, an aggregation module is developed to efficiently combine the relative features across both spatial and temporal domains. The objective of this operation is to update the feature maps $\Hmat^{(l+1)}$, which have the same dimensions as the input $\Hmat^{(l)}$, using the learned dynamic graph-structured features to capture non-local dependencies and improve video reconstruction. {The aggregation process can be viewed as information interaction at each iteration, while the graph typically requires $L$ iterations to update features.}

Towards this end, we define a generic aggregation operation for updating each node $\hv_i^{(l)}$ at the $l$-th iteration step as:
\begin{eqnarray}
\hv_{i}^{(l+1)} = \textstyle\frac{1}{\mathcal{C}(\hv_i^{(l)})} \sum_{b=1}^{B} \sum_{ j \in \mathcal{N}({\hv_{i}^{(l)}})} \mathcal{R}(\hv_{i}^{(l)}, \hv_{j,b}^{(l)}) f(\hv_{j,b}^{(l)}),
\\
 {\mathcal{C}(\hv_i^{(l)})}= \textstyle\sum_{b=1}^{B} \sum_{ j \in \mathcal{N}({\hv_{i}^{(l)}})} \mathcal{R} (\hv_{i}^{(l)}, \hv_{j,b}^{(l)}),
\end{eqnarray}
where $\hv_i^{(l+1)}$ is the corresponding updated node feature; $\mathcal{N}({\hv_{i}^{(l)}})$ represents the neighboring node set of $\hv_{i}^{(l)}$, containing the dynamic sampled neighbors $\hv_{j,b}^l$ at the $b^{th}$ frame and the shared global graph nodes $\hv^{global}$; $\mathcal{R}\left(\hv_{i}^{(l)},\hv_{j,b}^{(l)}\right)$ represents the correlation between two adjacent nodes, performing as a weight of corresponding node feature $f(\hv_{j,b}^{(l)})$; ${\mathcal{C}(\hv_i^{(l)})}$ is utilized for normalization. Specifically, there could be different choices for the pairwise function $\mathcal{R}$ for measuring the relative relationship between two nodes. In this paper, the embedded Gaussian version is utilized to compute node relationships as:
\begin{equation}
\textstyle \mathcal{R}\left(\mathbf{h}_{i}^{(l)}, \mathbf{h}_{j,b}^{(l)}\right)= e^{f\left(\mathbf{h}_{i}^{(l)}\right) f\left(\mathbf{h}_{j,b}^{(l)}\right)} \wv_b,
\end{equation}
where the function $f$ is composed of CNNs for transforming the input node features to another representation space to calculate nodes' similarities. Moreover, the frame-wise weight $\wv_b$ is a learnable parameter to `re-weight' the node relationships of the $b^{th}$ frame, as we assume the node features of some frames (might be the neighboring frames) are more important for updating node $\mathbf{h}_{i}^{(l)}$ than other frames. For ease of comprehension, the aggregation operation is similar to the self-attention mechanism in \cite{Cheng20ECCV_BIRNAT, Transformer}, but performed on the dynamically selected relative nodes. From this perspective, our aggregation can be viewed as an efficient variety of self-attention mechanisms, more appropriate for video processing.

{Up to now, the features of dynamical graph nodes can be adaptively updated by aggregating the relative information regardless of the spatial or temporal distance. {The non-locally aggregated information facilitates extracting the residual complementary features, which are combined with the initial prediction and subsequently fed into a 4-layer 3D CNN $\mathcal{F}_d$ to reconstruct the final fine-grained video $\tilde \Xmat $ as:}
\begin{equation} \label{Eq:recon}
\tilde \Xmat = \mathcal{F}_d(\Hmat^{(L)}+\Xmat^{Init}).
\end{equation}}

\noindent{\textbf{Optimization:}
At the training stage, we optimize our proposed model with mean square error (MSE) loss, \ie,
\begin{equation} \label{Eq:loss}
\textstyle \mathcal{L}_{mse} = \frac{1}{BHW} \sum_{b=1}^{B}||\tilde \Xmat_b - \Xmat_b ||^2_2,
\end{equation}
where $\tilde \Xmat_b$ is the $b^{th}$ frame of our final reconstructed video, and $\Xmat_b$ is the ground truth.
In order to efficiently obtain the initial prediction, we jointly train MadyGraph with loss:
\begin{equation} \label{Eq:loss_final}
\mathcal{L} = \textstyle \mathcal{L}_{mse} + \frac{1}{BHW} \sum_{b=1}^{B}||\Xmat^{Init}_b - \Xmat_b ||^2_2.
\end{equation}}

\section{Experiments}

This section presents the performance comparison of our proposed video reconstruction method, MadyGraph, {with several alternatives on both grayscale and color simulation datasets, as well as the real datasets.} The evaluation of different video gain SCI reconstruction methods on simulated datasets is based on the peak-signal-to-noise ratio (PSNR) and structured similarity index metrics (SSIM). We also demonstrate the ablation study to investigate the importance of each component in our MadyGraph.

\subsection{Datasets and Implement Details}

\noindent{\textbf{Simulation Video Datasets:}} 
We choose \texttt{DAVIS2017}~\cite{Pont-TusetPCASG17} as the training set for our experiments, which contains 90 different scenes of total 6208 frames with two resolutions: $480\times894$ and $1080\times1920$. 
During training, we perform data augmentation on  \texttt{DAVIS2017}~\cite{Pont-TusetPCASG17} using random horizontal flipping, random scaling, and random cropping following~\cite{wang2022spatial}. Randomly cropped patch cubes from the original scenes in \texttt{DAVIS2017} are synthesized as training data. Following the CACTI imaging process, a series of measurements are generated.
The evaluation experiments are implemented on the following datasets: 1) A widely used benchmark consists of six grayscale simulation datasets, including \texttt{Kobe, Runner, Drop, Traffic}~\cite{Liu18TPAMI}, \texttt{Aerial} and \texttt{Vehicle}~\cite{Yuan20PnPSCI}, each with a size of $256\times256\times8$. Following the setting in previous works~\cite{wang2022spatial, Cheng20ECCV_BIRNAT}, eight (B = 8) sequential frames of size $256\times256$ are modulated by the shifting binary masks and then collapsed into measurements Y for training.
2) Furthermore, we evaluate our model on six benchmark color simulation datasets including \texttt{Beauty}, \texttt{Bosphorus}, \texttt{Jockey}, \texttt{Runner}, \texttt{ShakeNDry} and \texttt{Traffic}~\cite{PnP_SCI_arxiv2021}, each with a size of $512\times512\times3\times32$. Here, the third dimension 3 represents the RGB channels. Similar to the grayscale video, we compress the video with a compression rate of B = 8, resulting in 4 measurements for each dataset. To generate the measurement for training, we use the same shifting binary masks as used in~\cite{PnP_SCI_arxiv2021} to modulate mosaic videos of size $512\times512$.

\noindent{\textbf{Real Video Datasets:}} For the grayscale real data, we choose three gray-scale scenes from two real SCI systems for testing, including \texttt{Wheel}~\cite{Patrick13OE} with the size of $256\times256\times14$, \texttt{Domino} and \texttt{Water Balloon}~\cite{Qiao2020_APLP} with the size of $512\times512\times10$. For the color real data, we use \texttt{Hammer} video data~\cite{Yuan14CVPR} with a spatial resolution of $512 \times 512$ and compression rate $B=22$. 

\begin{table}[t!]
\centering
\caption{Network architecture of the initial predictor.}
\resizebox{0.48\textwidth}{!}{
\begin{tabular}{c|c|c|c|c}
\toprule[1pt]
    \textbf{Module} & \textbf{Operation} & \textbf{Kernel} & \textbf{Stride}  & \textbf{Output Size} 
\\ \midrule
\multirow{4}{*}{\textbf{3D CNN block1}} 
& 3D conv. & $5\times 5\times 5$ &1 & $256\times256\times16\times8$\\
& 3D conv. & $3\times 3\times 3$ &1 & $256\times256\times32\times8$\\
& 3D conv. & $1\times 1\times 1$ &1 & $256\times256\times32\times8$\\
& 3D conv. & $3\times 3\times 3$ &2 & $128\times128\times64\times8$\\\midrule
\multirow{1}{*}{\textbf{Resblock1}} 
& $resblock$ & $\begin{bmatrix}
    3\times 3\times 3\\
    1\times 1\times 1\\
    3\times 3\times 3\\
\end{bmatrix}\times 3$ &1 & $128\times128\times64\times8$\\\midrule
\multirow{2}{*}{\textbf{3D CNN block2}}
& 3D conv. & $3\times 3\times 3$ &1 & $128\times128\times64\times8$\\
& 3D conv. & $1\times 1\times 1$ &1 & $128\times128\times64\times8$\\\midrule
\multirow{1}{*}{\textbf{Resblock2}} 
& $resblock$ & $\begin{bmatrix}
    3\times 3\times 3\\
    1\times 1\times 1\\
    3\times 3\times 3\\
\end{bmatrix}\times 3$ &1 & $128\times128\times64\times8$\\\midrule
\multirow{6}{*}{\textbf{3D CNN block3}} 
& 3D conv. & $3\times 3\times 3$ &1 & $128\times128\times64\times8$\\
& 3D conv. & $1\times 1\times 1$ &1 & $128\times128\times64\times8$\\
& 3D deconv. & $3\times 3\times 3$ &2 & $256\times256\times 32\times8$\\
& 3D conv. & $3\times 3\times 3$ &1 & $256\times256\times 16\times8$\\
& 3D conv. & $1\times 1\times 1$ &1 & $256\times256\times 16\times8$\\
& 3D conv. & $3\times 3\times 3$ &1 & $256\times256\times 1\times8$\\
\bottomrule
\end{tabular}}
\label{table:basenet}
\end{table}

\begin{table}[t!]
\centering
\caption{Network architecture of the proposed MadyGraph.}
\resizebox{0.48\textwidth}{!}{
\begin{tabular}{c|c|c|c|c|c}
\toprule[1pt]
    \textbf{Module} & \multicolumn{2}{c|}{\textbf{Operation}} & \textbf{Kernel} & \textbf{Stride}  & \textbf{Output Size} 
\\ \midrule
\multirow{4}{*}{\textbf{Feature Extraction}  $\mathcal{F}_{e}$} 
& \multicolumn{2}{c|}{3D conv.} & $5\times 5\times 5$ &1 & $256\times256\times16\times8$\\
& \multicolumn{2}{c|}{3D conv.} & $3\times 3\times3$ &1 & $256\times256\times32\times8$\\
& \multicolumn{2}{c|}{3D conv.} & $1\times 1\times 1$ &1 & $256\times256\times64\times8$\\
& \multicolumn{2}{c|}{3D conv.} & $3\times 3\times 3$ &1 & $256\times256\times64\times8$\\\midrule
\multirow{5}{*}{\textbf{Dynamic Graph}} 
& \multicolumn{2}{c|}{\textbf{Resblock}} & $\begin{bmatrix}
    3\times 3\times 3\\
    1\times 1\times 1\\
    3\times 3\times 3\\
\end{bmatrix}\times 3$ &1 & $256\times256\times64\times8$\\\cmidrule{2-6}
&\multirow{1}{*}{\textbf{ {Scale 1}}} 
 & 3D conv. & $3\times3\times1$ &1 & $256\times256\times27\times8$\\\cmidrule{2-6}
&\multirow{1}{*}{\textbf{ {Scale 2}}} 
 & 3D conv. & $3\times3\times1$ &7 & $256\times256\times27\times8$\\\cmidrule{2-6}
&\multirow{1}{*}{\textbf{ {Scale 3}}} 
 & 3D conv. & $3\times3\times1$ &13 & $256\times256\times27\times8$\\\cmidrule{2-6}
& \multicolumn{2}{c|}{3D conv.} & $3\times3\times 3$ &1 & $256\times256\times64\times8$\\
\midrule
\multirow{4}{*}{\textbf{Reconstruction}  $\mathcal{F}_{d}$} 
& \multicolumn{2}{c|}{3D conv.} & $3\times 3\times 3$ &1 & $256\times256\times32\times8$\\
& \multicolumn{2}{c|}{3D conv.} & $1\times1\times1$ &1 & $256\times256\times32\times8$\\
& \multicolumn{2}{c|}{3D conv.} & $3\times3\times3$ &1 & $256\times256\times16\times8$\\
& \multicolumn{2}{c|}{3D conv.} & $1\times1\times1$ &1 & $256\times256\times1\times8$\\
\bottomrule
\end{tabular}}
\label{table:setting}
\end{table}

\noindent{\textbf{Counterparts and Evaluation Metrics:}}
The proposed MadyGraph is compared with several methods including model-based methods GAP-TV~\cite{Yuan16ICIP_GAP} and DeSCI~\cite{Liu18TPAMI}, plug-and-play method PnP-FFDNet~\cite{Yuan20PnPSCI}, and deep learning based methods E2E-CNN~\cite{Qiao2020_APLP}, BIRNAT~\cite{Cheng20ECCV_BIRNAT}, RevSCI~\cite{Cheng2021_CVPR_ReverSCI}, DDUN~\cite{wu2021SCI3D}, STFormer~\cite{wang2022spatial}, EfficientSCI~\cite{Wang_2023_CVPR}. For simulation data, both peak-signal-to-noise ratio (PSNR) and structural similarity (SSIM)~\cite{Wang04imagequality} are employed as metrics to quantitatively evaluate the reconstruction quality.
Moreover, to measure the practicality of the methods for real-time applications, we report the inference time for reconstructing videos during the testing stage.

\noindent{\textbf{Implement Details:}}
{We choose 3 dilation rates 1, 7, and 13 (note that they are all prime numbers) for the cross-scale node sampling in the dynamic graph.}
For each latent node, we sample $K=9$ neighboring nodes of each scale. The patch size of each node is set as $p=4$. The number of global nodes is 16 ($4\times4$). 
EfficientSCI-B~\cite{Wang_2023_CVPR} and STFormer~\cite{wang2022spatial} are employed as the backbone models.
The Adam optimizer~\cite{KingmaB14} is employed for optimization with the initial learning rate of {$2e^{-4}$}. All experiments are implemented in PyTorch running on an RTX 8000 GPU.

\noindent\textbf{Architecture of the initial predictor:}
{We list the detailed architecture of the lightweight initial predictor in Table~\ref{table:basenet}, corresponding to the top row of Fig~\ref{fig:model}, which is only utilized at the training stage.}

\begin{table*}[htbp!]
  \caption{The average results of PSNR in dB (left entry in each cell), SSIM (right entry in each cell), and running time per measurement/shot in seconds by different algorithms on the six grayscale {\em benchmark} data. The best results are in \textbf{bold}.}
\begin{center}
  \resizebox{1\textwidth}{!}{
  \begin{tabular}{c|c|c|c|c|c|c|c|c}
    \toprule
    Algorithm & \texttt{Kobe} & \texttt{Traffic} &\texttt{Runner} &\texttt{Drop} &\texttt{Aerial} &\texttt{Vehicle} &Average & Time\\
    \midrule\midrule
    GAP-TV~\cite{Yuan16ICIP_GAP} &26.45, 0.845 &20.89, 0.715  &28.81, 0.909 &34.74, 0.970 &25.05, 0.828  &24.82, 0.838 &26.79, 0.858 &4.20\\\midrule
    PnP-FFDNet~\cite{Yuan20PnPSCI} & 30.50, 0.926 & 24.18, 0.828 & 32.15, 0.933 & 40.70, 0.989 & 25.27, 0.829 & 25.42, 0.849 & 29.70, 0.892 & 3.0\\\midrule
    E2E-CNN~\cite{Qiao2020_APLP} & 29.02, 0.861 & 23.45, 0.838 & 34.43, 0.958 & 36.77, 0.974 & 27.52, 0.882 & 26.40, 0.886 & 29.26, 0.900 & \textbf{0.023}\\ \midrule   
    DeSCI~\cite{Liu18TPAMI}  &{33.25}, {0.952}  &{28.72}, 0.925  &{38.76}, {0.969}  &{43.22},  {0.993}  &25.33,  0.860  &27.04,  0.909  &32.72, 0.935 &6180\\\midrule
    BIRNAT~\cite{Cheng20ECCV_BIRNAT} &32.71, 0.950 & 29.33, {0.942} &38.70, 0.976 &42.28, 0.992 &{28.99}, {0.927} &{27.84}, {0.927} &{33.31}, {0.951} &0.16\\\midrule
    GAP-net-Unet-S12~\cite{Meng_GAPnet_arxiv2020} &32.09, 0.944 &28.19, 0.929 &38.12, 0.975 &42.02, 0.992 &27.83, 0.931 &28.88, 0.914 &32.86, 0.947 &0.03\\\midrule
    MetaSCI~\cite{wang2021metasci}	&30.12, 0.907 &26.95, 0.888 &37.02, 0.967 &40.61, 0.985 &27.33, 0.906 &28.31, 0.904 &31.72, 0.926 &0.03\\\midrule
    RevSCI~\cite{Cheng2021_CVPR_ReverSCI} &{33.72}, {0.957} &{30.02}, {0.949} &{39.40}, {0.977} &{42.93}, {0.992} &{29.35}, {0.924} &{28.12}, {0.937}  &{33.92}, {0.956} &{0.19}\\\midrule
    DDUN~\cite{wu2021SCI3D} &35.02, 0.968  &31.78, 0.964 &40.91, 0.982 &44.49, 0.994 &30.58, 0.940 &29.36, 0.955 &35.36, 0.967 &1.35\\\midrule
    STFormer~\cite{wang2022spatial}	&35.50, 0.973 &32.11, 0.967 &42.66, 0.988 &44.55,0.994 &31.48, 0.952 &31.15, 0.972 &36.24, 0.974 &0.92\\\midrule
    EfficientSCI-B~\cite{Wang_2023_CVPR} &35.76, 0.974 &32.30, 0.968 &43.05, {0.988} &45.18, 0.995 &31.50, 0.953 &31.13, 0.971 &36.48, 0.975 &0.31 \\\midrule\midrule
    \textbf{MadyGraph} &\textbf{37.30, 0.981} &\textbf{33.13, 0.973} &\textbf{44.13, 0.990}  &\textbf{45.95,  0.996}  &\textbf{32.12, 0.959} &\textbf{31.82, 0.975} 
 &\textbf{37.41, 0.979} &0.41\\\bottomrule
  \end{tabular}}
  \end{center}
  \label{Table:sim-grayscale}
\end{table*}

\noindent\textbf{Architecture of the graph module:}
{We list the detailed network architectures of the graph module in Table~\ref{table:setting}, corresponding to the bottom row of Fig~\ref{fig:model}. The graph module mainly contains four components:
i) A feature extraction block $\mathcal{F}_{e}$, a 4-layer 3D CNN, encodes the re-masked images into high-dimensional (4D) features.
ii) {An optical flow extraction network, which is a lightweight pre-trained model. The detailed architecture and implementation can be found in }\url{https://github.com/twhui/LiteFlowNet}.
iii) A non-local dynamic graph neural network is proposed to aggregate the cross-scale dynamically selected features for modeling pixel-wise spatial-temporal correlations.
iv) A 4-layer 3D CNN module $\mathcal{F}_{d}$ is employed to reduce the channel and achieve the final enhanced reconstructed video.}

\subsection{Results on Simulation Datasets}

\begin{figure*}[!th]
\centering
\includegraphics[width=1.0\textwidth]{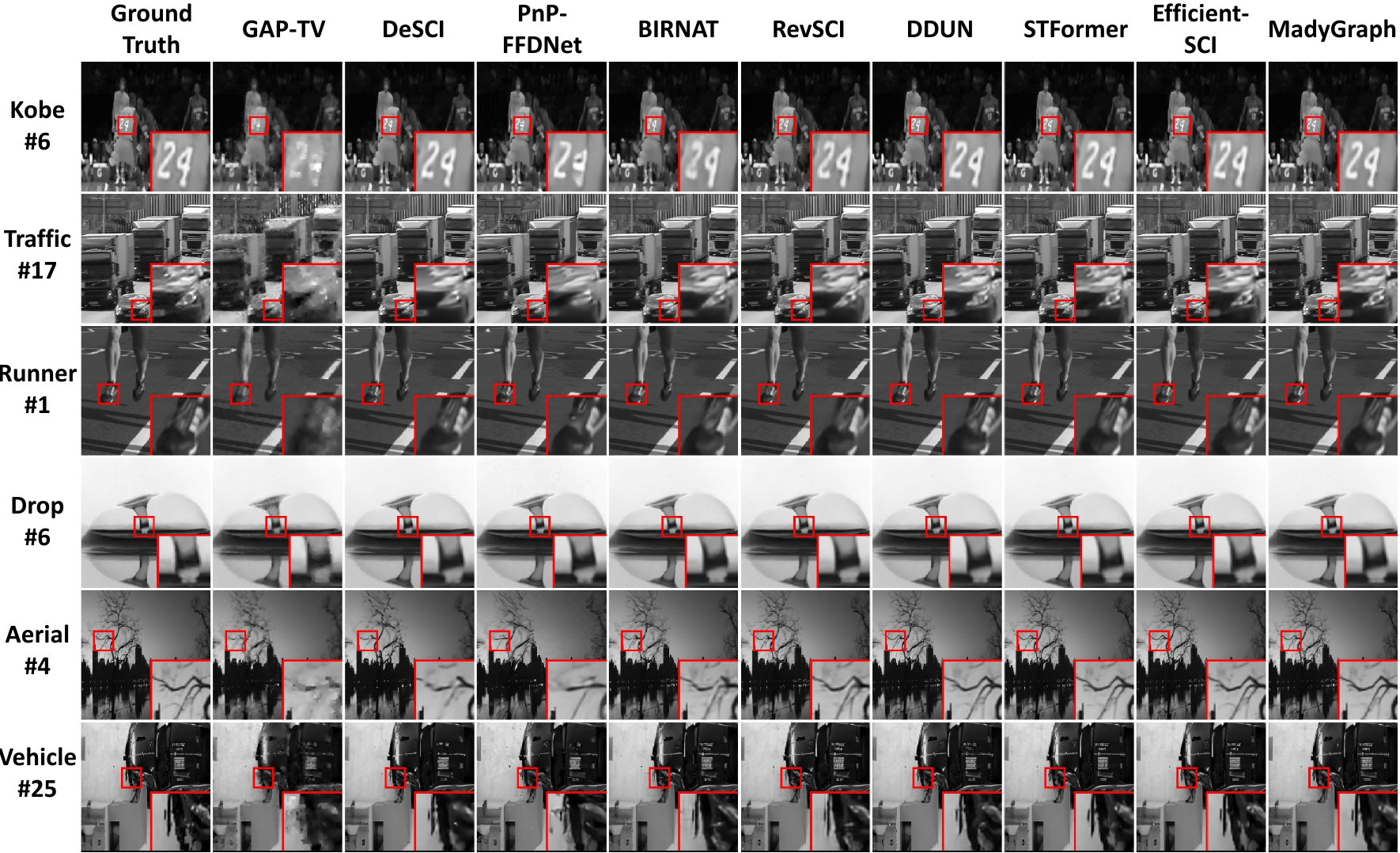}
\caption{\label{fig:sim-grayscale} {Selected reconstructed frames of six benchmark grayscale simulation data of different methods.}}
\end{figure*}

\begin{figure}[!t]
\centering
\includegraphics[width=0.5\textwidth]{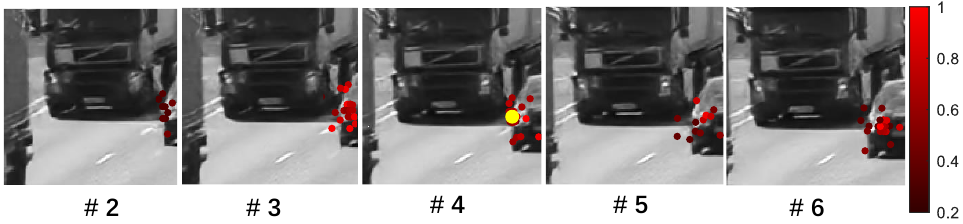}
\caption{\label{fig:dynamic} {{Example of the dynamic nodes learned by MadyGraph on \texttt{Traffic}. The yellow point in the middle Frame \#4 is the central point, and the red points denote the dynamically selected neighbors with high aggregation weights larger than 0.2.}} 
}
\end{figure}

\subsubsection{Grayscale Simulation Video}

The comparison results of six widely used benchmark grayscale simulation data are given in Table~\ref{Table:sim-grayscale}, comparing with different algorithms, \ie,  GAP-TV~\cite{Yuan16ICIP_GAP}, PnP-FFDNet~\cite{Yuan20PnPSCI}, E2E-CNN~\cite{Qiao2020_APLP}, DeSCI~\cite{Liu18TPAMI}, BIRNAT~\cite{Cheng20ECCV_BIRNAT}, GAP-net-Unet-S12~\cite{Meng_GAPnet_arxiv2020}, MetaSCI~\cite{wang2021metasci},
RevSCI~\cite{Cheng2021_CVPR_ReverSCI}, DDUN~\cite{wu2021SCI3D}, STFormer~\cite{wang2022spatial}, and SOTA method EfficientSCI~\cite{Wang_2023_CVPR}. Fig.~\ref{fig:sim-grayscale} {visualizes} selected reconstructed frames of our method on the six datasets, as well as the other comparison methods. The observations from Table~\ref{Table:sim-grayscale} and Fig.~\ref{fig:sim-grayscale} are listed as follows.

$i$) MadyGraph consistently outperforms the previous comparison methods on six benchmark data. To be specific, MadyGraph surpasses the SOTA method EfficientSCI on \texttt{Kobe}, \texttt{Traffic}, \texttt{Runner}, \texttt{Drop}, \texttt{Aerial}, and \texttt{Vehicle} by \{1.54, 0.83, 1.08, 0.77, 0.62, 0.69\} on the metric PSNR, respectively. It can be found that MadyGraph can provide superior {performances} on videos with faster-moving action, such as running or playing basketball. It might own to the non-local spatial-temporal correlation modeling by our dynamic graph, which can flexibly aggregate the most relative contextual information without much redundancy. For the scenes with complicated backgrounds or moving objects such as \texttt{Kobe} and \texttt{Aerial}, the non-local similarity-based method DeSCI tends to degrade, and correlation-based deep learning methods such as STFormer and our MadyGraph can provide consistent robustness due to the strong representation capability for long-range dependencies.

$ii$) {Considering the reconstruction running time as shown in the last column of Table~\ref{Table:sim-grayscale}, our proposed method strikes a balance between running performance and reconstruction quality, achieving competitive results in both aspects. In specific, our MadyGraph achieves a high reconstruction quality of over 37 dB, while maintaining a fast running time of 0.41s. These results outperform most current reconstruction algorithms in terms of running speed and reconstruction quality. Although GAP-net-Unet, MetaSCI, and BIRNAT have higher running speeds, their reconstruction quality is relatively poor, with an average PSNR value of less than 34 dB. {Compared with the Transformer-based method STFormer, which also aims at modeling long-range dependencies, our MadyGragh delivers twice the inference speed and significantly better reconstruction quality, boasting an average PSNR improvement of 1.17dB.} This contributes to the sparse and flexible sampling strategy of graph nodes, which is lightweight and effective in alleviating the severe computational problems caused by the self-attention mechanism.}

$iii$) In terms of visualization quality, it can be seen from Fig.~\ref{fig:sim-grayscale} that our MadyGraph is able to clearly reconstruct the details. Especially for high-speed scenarios with fast motions (such as \texttt{Kobe}),  MadyGraph has excellent performance for reconstructing the actions (such as playing basketball). We contribute it to the powerful capacity of non-Euclidean distance modeling and long-term dependencies exploring. MadyGraph could adaptively capture the relations between basketball and the player with prior knowledge from optical flow. {By analyzing the visualization results, we can observe that our proposed method successfully recovers clear edges of tree trunks in the \texttt{Aerial} data, as well as the car lamp on the cars in the \texttt{Traffic} data.} {Previous approaches often result in over-smoothing and blurred periphery. In contrast, our MadyGraph offers sharper edges and more distinct contours due to its strong representation capability.}

Furthermore, to showcase how the proposed model finds related clues to update the representation for reconstructing videos, we visualize the intrinsic dynamic relationships in Fig.~\ref{fig:dynamic}. Intuitively, we plot the relations at the central position (\ie, the yellow point) with the dynamically selected nodes (\ie, the red points) with various shades of color. {As the car moves forward over time}, the focal point resides at the headlight of a moving vehicle, becoming increasingly prominent in the subsequent frames.
Our observations are as follows: 1) The majority of dynamically selected neighboring nodes are situated on the car lamp or in its surrounding vicinity.
2) Over time, the selected neighbors tend to disperse around the lamp while also concentrating on the right side of the image, aligning with the movement of the object.
3) In frame \#1, the aggregation weights of neighboring nodes are below 0.2, in accordance to that the lamp is not visible in the image. This indicates that our proposed model can dynamically capture meaningful and logical correlations regardless of the distance across space and time in the video.

\begin{table*}[t!]
\caption{\small The average results of PSNR in dB (left entry) and SSIM (right entry) and running time per measurement/shot in seconds by different algorithms on six color simulation datasets. The best results are in {\bf bold}.}
\begin{threeparttable}
\begin{center}
\resizebox{1.0\textwidth}{!}{
\begin{tabular}{c|c|c|c|c|c|c|c|c}
\toprule[1pt]
Algorithm &\texttt{Beauty} &\texttt{Bosphorus} &\texttt{Jockey}  &\texttt{Runner} &\texttt{ShakeNDry} &\texttt{Traffic} &Average &Time 
\\ \midrule\midrule
{GAP-TV}~\cite{Yuan16ICIP_GAP}  &33.08, 0.9639 &29.48, 0.9144 &29.48, 0.8874   &29.10, 0.8780  &29.59, 0.8928 &19.84, 0.6448   &28.46, 0.8635 &10.8  \\  \midrule
{DeSCI}~\cite{Liu18TPAMI}  &34.66, 0.9711 &32.88, 0.9518  &34.14, 0.9382 &36.16, 0.9489  &30.94  , 0.9049 &24.62, 0.8387  &32.23, 0.9256 &92640  \\  \midrule
{PnP-FFDNet}~\cite{Yuan20PnPSCI}  &33.21, 0.9629 &28.43, 0.9046 &32.30, 0.9182 &30.83, 0.8875 &27.87, 0.8606 &21.03, 0.7113  &28.93, 0.8742 &25.8\\ \midrule
{PnP-FastDVDnet-gray~\cite{PnP_SCI_arxiv2021}} &33.01, 0.9628 &33.01, 0.9628 &33.51, 0.9279 &32.82, 0.9004 &29.92, 0.8920 &22.81, 0.7764 &30.50, 0.8989 &52.2 \\ \midrule
{PnP-FastDVDnet-color~\cite{PnP_SCI_arxiv2021}} &{35.27}, {0.9719} &{37.24}, {0.9781} &{35.63}, {0.9495} &{38.22}, {0.9648} &{33.71}, {0.9685} &{27.49}, {0.9147} &{34.60}, {0.9546} &57\\ \midrule
BIRNAT-color~\cite{cheng2022recurrent} &{36.08}, {0.9750} &{38.30}, {0.9817} &{36.51}, {0.9561} &{39.65}, {0.9728} &{34.26}, {0.9505} &{28.03}, {0.9151} &{35.47}, {0.9585} &\textbf{0.98}\\ \midrule
STFormer*~\cite{wang2022spatial} &36.88, 0.977 &40.57, 0.989 &38.10, 0.965 &42.43, 0.984 &34.98, 0.957 &29.98, 0.941 &37.16, 0.969 &1.95\\ \midrule\midrule
\textbf{MadyGraph} &\textbf{37.13, 0.978} &\textbf{40.74, 0.989} &\textbf{38.43, 0.967} &\textbf{42.58, 0.985} &\textbf{36.01, 0.965} &\textbf{30.38, 0.945} &\textbf{37.55, 0.971} & 2.05\\\bottomrule
\end{tabular}}
\end{center}
\begin{tablenotes}
\item[* We use the official weights from https://github.com/ucaswangls/STFormer to generate the results and then calculate the evaluation metrics on the RGB videos.]
\end{tablenotes}
\end{threeparttable}
\label{Table:sim-color}
\end{table*}

\begin{figure*}[!th]
\centering
\includegraphics[width=0.9\textwidth]{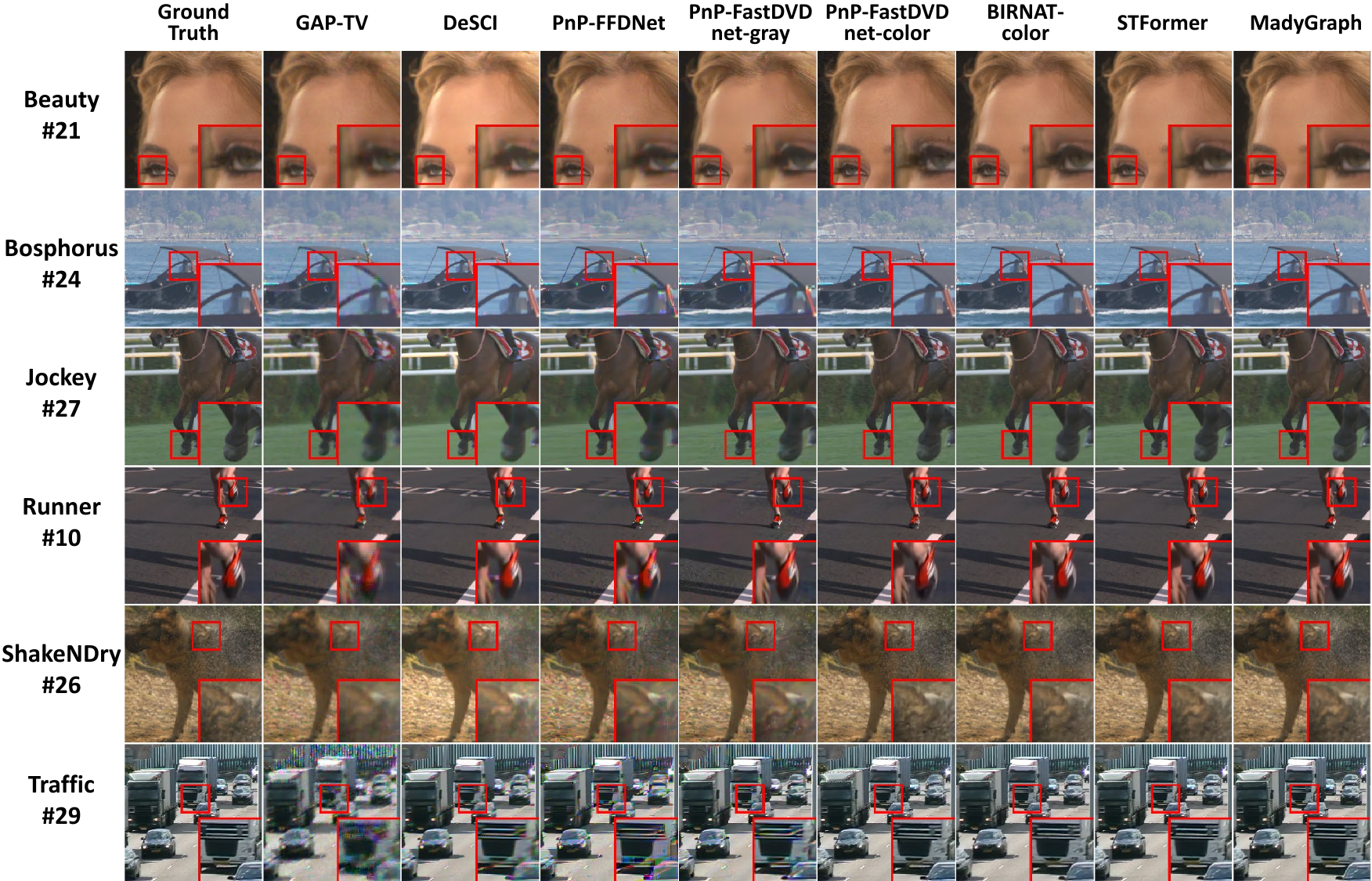}
\caption{\label{fig:sim-color} {{Selected reconstructed frames of six benchmark color simulation data of different methods.}} 
}
\end{figure*}

\subsubsection{Color Simulation Video} 
Table~\ref{Table:sim-color} presents the quantitative results of our proposed method on six widely used benchmark color simulation datasets, along with other comparison algorithms.
Due to the inflexibility of DDUN~\cite{wu2021SCI3D} and RevSCI~\cite{Cheng2021_CVPR_ReverSCI} with regard to input size and masks, training a model with a spatial size of $512 \times 512$ requires a significant amount of memory or training time. Therefore, we limit our comparison to two iterative optimization-based algorithms, \ie, GAP-TV~\cite{Yuan16ICIP_GAP} and DeSCI~\cite{Liu18TPAMI}, three PnP algorithms, an end-to-end deep learning algorithm BIRNAT-color~\cite{Cheng20ECCV_BIRNAT}, and the SOTA method STFormer~\cite{wang2022spatial} on color simulation videos. It is worth noting that PnP methods are categorized into gray and color versions based on the use of grayscale or color denoisers. 
Selected reconstructed frames of our method and the other comparison methods on these six datasets are visually represented in Fig.~\ref{fig:sim-color}. The key observations of Table~\ref{Table:sim-color} and Fig.~\ref{fig:sim-color} can be summarized as follows:

$i$) According to the quantitative results, the PSNR value of MadyGraph is 37.55 dB, surpassing the PSNR values of all other compared algorithms. Taking the results on the image \texttt{ShakeNDry} for example, which contains minor shaking motions of fur in the foreground and tangled weeds in the background, our proposed MadyGraph outperforms the previous state-of-the-art method STFormer by 1.03dB on PSNR. This indicates that our model can effectively capture both the local dynamic details and the complex background of a scene. Overall, the superior results verify that MadyGraph is also effective in accurately reconstructing color high-speed scenes.

$ii$) With regard to the running time for reconstructing videos, our proposed MadyGraph achieves higher reconstruction quality with competitive speed. {Specifically, the iterative optimization-based DeSCI algorithm requires over 24 hours for reconstruction.} Three PnP-based algorithms take more than 20 seconds. Two recently developed deep learning-based reconstruction algorithms, BIRNAT-color and STFormer, have shown promising results with a fast runtime of only 2 seconds, which is over 10 times faster than previous PnP methods. {Consistent with other deep learning methods with fast inference speed, our proposed model showcases good real-time performance.}

$iii$) Based on the visualization results as shown in Fig.~\ref{fig:sim-color}, our proposed method proves to be effective in recovering sharp edges and capturing detailed complex textures. {Taking the \texttt{Runner} dataset as an example, MadyGraph can provide clearer shoe textures with fewer artifacts than the other competing methods. However, optimization-based methods including GAP-TV and DeSCI, fail to jointly optimize SCI and demosaicing, leading to suboptimal color video SCI reconstructions. Although PnP-FFDNet-color and PnP-FastDVD-color methods perform better in detail preservation, they still suffer from blurred edges.}

\begin{figure}[!t]
\centering
\includegraphics[width=0.5\textwidth]{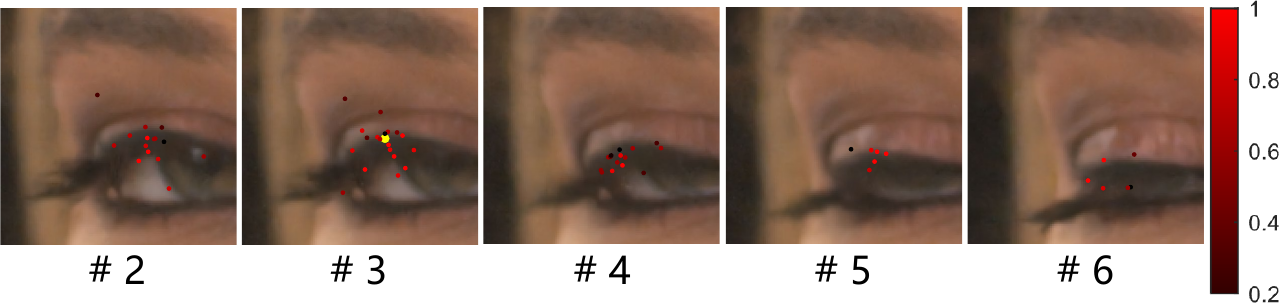}
\caption{\label{fig:dynamic-color} {{Example of the dynamic nodes learned by MadyGraph on color SCI video \texttt{Beauty}. The yellow point in the middle Frame \#3 is the central point, and the red points denote the dynamically selected neighbors with high aggregation weights larger than 0.2.}} }
\end{figure}

\textbf{{Visualization of the dynamic correlations on color videos:} }
{The visualization in Fig.~\ref{fig:dynamic-color} showcases the intrinsic dynamic relationships investigated by the graph during a woman blinking her eye in the video dataset \texttt{Beauty}. The yellow point at the center of frame \#3 represents the basement point, while the red points surrounding it are selected as the related neighbor nodes of the basement point.
It is evident from the visualization that the majority of the selected relative nodes are adaptively flexible and mostly located on the woman’s eyelashes. Additionally, as the eye gradually blinks shown in the following frames, the relative selected nodes become sparser and move downward along with the eyelash. This observation demonstrates that the proposed model is capable of capturing meaningful dynamical correlations, irrespective of the spatial and temporal distances involved in the color video, and well handling the fast movements.}

\begin{table}[t!]
  \caption{{{Computational complexity and average reconstruction quality on six benchmark simulation data for ablation study of MadyGraph. MAC means Multiply Accumulate.}}}
\scriptsize
\begin{center}
  \resizebox{0.48\textwidth}{!}{
  \begin{tabular}{cccc|c|c|c|c|c|c}
    \toprule
    DW &CS &MA &GK &PSNR &SSIM  &MACs &Parameters &Time &{Memory}\\ 
    & & & & & & ($\times10^{11}$) & ($\times10^{6}$) & &{(MB)}\\ \midrule
    $\times$ &$\times$  &$\times$ &$\times$ &36.57 &0.975 &2.53 &6.56 &0.33 &{1050}
    \\ \midrule   
    \checkmark &$\times$  &$\times$  &$\times$
    &37.08 &0.977 &3.21 &7.17 &0.35 &{2474}\\ \midrule
    \checkmark &\checkmark &$\times$ &$\times$ &37.21 &0.978  &3.38 &7.29 &0.36 &{3290} \\ \midrule  
    \checkmark &\checkmark &\checkmark &$\times$
    &37.30 &0.978  &8.55 &7.62 &0.40 &{3775}\\ \midrule  
    \checkmark &\checkmark &\checkmark &\checkmark
    &37.41 &0.979  &8.75 &7.71 &0.41 &{4432}\\
    \bottomrule
  \end{tabular}}
  \end{center}
  \label{Table:ablation}
\end{table}

\subsection{{Ablation Study}}

{To quantitatively verify the contributions of each module in the MadyGraph, we add the partial components step by step with average results shown in Table~\ref{Table:ablation} on the grayscale simulated data, {specifically corresponding to:}

\noindent\textbf{1) Dynamic walk (DW) sampling strategy}: The dynamic walk sampling could benefit from high adaptability and non-local spatial-temporal dependencies, leading to a significant performance increment by 0.51dB on PSNR compared with initial sampling at the fixed position. Furthermore, the computation cost is well controlled with minor parameters overhead by sparsely modeling non-local dependencies.

\noindent\textbf{2) Cross-scale (CS) node sampling mechanism}: In contrast to performing graph aggregation at just one scale, our cross-scale node sampling scheme brings about a performance enhancement of 0.13 dB. This enlargement enables multi-level representations and naturally increases the receptive field size. Additionally, this parallel architecture results in a reasonable increase of parameters ($0.12 \times 10^6$) and inference time (0.01s), which are both acceptable.

\noindent\textbf{3) Motion-aware (MA) information}: The prior knowledge of motion information (achieved by motion-aware dynamic walks) is incorporated to help the model find the positions of related nodes according to the motion over time in a video. It can be seen that the motion-aware mechanism achieves a performance improvement (0.09 dB) by leveraging motion knowledge for better video representation. Although the optical flow predictor leads to increased parameters, it is worth noting that the increased parameters need no training as the optical flow predictor has been already pre-trained.

\noindent\textbf{4) Global knowledge (GK) integration}: The higher-level knowledge embedded in global graph nodes not only helps to identify and connect with relevant low-level nodes but also gains a holistic view of the entire video. It can be seen that the global knowledge achieves a performance improvement of 0.11 dB for video representation. Note that optical flow is utilized to highlight the foreground moving objects, contributing to each node’s feature update.

\begin{table}[!t]
\caption{{PSNR and memory usage of different sampling strategies.}}
\centering
\begin{tabular}{c|c|c}
\bottomrule
{Sampling Method} & {PSNR} & {Memory (GB)} \\ \midrule\midrule
Fixed-stride & 36.57  & 2.0                           \\ \midrule
Sparse Attention-based  & 36.64         &1.1                           \\  \midrule
Entropy-based        & 36.94         &8.1                           \\ \midrule
MSE-based            & 36.89         & 8.1                           \\ \midrule
Self-Attention-based  & 37.11         &8.2                           \\ \midrule\midrule
our dynamic graph  & 37.41         & 2.4                           \\ \midrule
\end{tabular}
\label{table:entropy}
\end{table}

{We have also conducted experiments on grayscale simulated videos by replacing our dynamic sampling strategy with patch-level sorting methods for sampling, including fixed stride sampling, sparse attention~\cite{child2019generating}, entropy-based sampling, MSE-based sampling, and self-attention~\cite{Transformer}.
For fixed stride sampling, entropy-based, and MSE-based sampling approaches, we sample 27 neighboring nodes for each node, aligning with the number used in our proposed Graph module ($3\times 3$ neighboring nodes across 3 scales).
Unlike these methods, self-attention operates globally, and the sparsity pattern of sparse attention is data-driven. 
The PSNR results and memory usage of these different sampling strategies are presented in Table~\ref{table:entropy}.
As demonstrated in Table~\ref{table:entropy}, the proposed graph module not only achieves optimal performance but also maintains an appropriate level of memory usage. While fixed stride sampling and sparse attention having lower memory footprints, they yield comparatively poorer performance.}

{Overall, the comprehensive version of MadyGraph performs best, demonstrating its effectiveness and efficiency as a harmonious combination of its components. Various components of MadyGraph can work well together to produce superior performance with an appropriate memory usage.}

\subsection{Genralizability}

\begin{table*}[th!]
  \caption{The average results of PSNR in dB (left entry in each cell), SSIM (right entry in each cell), and running time per measurement/shot in seconds on the six grayscale {\em benchmark} data applied on different backbones.}
\begin{center}
  \resizebox{1\textwidth}{!}{
  \begin{tabular}{ccccccccc}
    \toprule
    Algorithm & \texttt{Kobe} & \texttt{Traffic} &\texttt{Runner} &\texttt{Drop} &\texttt{Aerial} &\texttt{Vehicle} &Average & Time\\
    \midrule
    GAP-TV~\cite{Yuan16ICIP_GAP} &26.45, 0.845 &20.89, 0.715  &28.81, 0.909 &34.74, 0.970 &25.05, 0.828  &24.82, 0.838 &26.79, 0.858 &4.20\\
    \rowcolor{mygray}GAP-TV + Graph-module &28.97, 0.911 &23.12, 0.823 &32.86, 0.955 &38.96, 0.987 &26.98, 0.882 &26.13, 0.887 &29.50, 0.908 &4.29\\ \midrule
    DeSCI~\cite{Liu18TPAMI}  &{33.25}, {0.952}  &{28.72}, 0.925  &{38.76}, {0.969}  &{43.22},  {0.993}  &25.33,  0.860  &27.04,  0.909  &32.72, 0.935 &6180\\
    \rowcolor{mygray}DeSCI + Graph-module &35.42, 0.967 &30.41, 0.947 &40.71, 0.977 &45.06, 0.994 &26.76, 0.899 &28.00, 0.929 &34.39, 0.952 &6180\\ \midrule
    BIRNAT~\cite{Cheng20ECCV_BIRNAT} &32.71, 0.950 & 29.33, {0.942} &38.70, 0.976 &42.28, 0.992 &{28.99}, {0.927} &{27.84}, {0.927} &{33.31}, {0.951} &0.16\\
    \rowcolor{mygray}BIRNAT + Graph-module &33.48, 0.944 &29.95, 0.946 &39.48, 0.978 &42.88, 0.992 &29.20, 0.921 &27.99, 0.931 &33.83, 0.952 &0.26\\ \midrule    RevSCI~\cite{Cheng2021_CVPR_ReverSCI} &{33.72}, {0.957} &{30.02}, {0.949} &{39.40}, {0.977} &{42.93}, {0.992} &{29.35}, {0.924} &{28.12}, {0.937}  &{33.92}, {0.956} &{0.19}\\
    \rowcolor{mygray}RevSCI + Graph-module &{34.34}, {0.962} &{30.59}, {0.956} &{40.34}, {0.981} &{43.56}, {0.993} &{29.56}, {0.928} &{28.20}, {0.940} &{34.43}, {0.960} &0.30\\ \midrule
    DDUN~\cite{wu2021SCI3D} &35.02, 0.968  &31.78, 0.964 &40.91, 0.982 &44.49, 0.994 &30.58, 0.940 &29.36, 0.955 &35.36, 0.967 &1.35\\
    \rowcolor{mygray}DDUN + Graph-module &{36.63}, {0.975} &{32.60}, {0.969} &{42.17}, {0.985} &{45.59}, {0.995} &{30.91}, {0.946} &{29.76}, {0.957} &{36.28}, {0.971} &1.45\\\midrule    STFormer~\cite{wang2022spatial}	&35.50, 0.973 &32.11, 0.967 &42.66, 0.988 &44.55, 0.994 &31.48, 0.952 &31.15, 0.972 &36.24, 0.974 &0.92\\
    \rowcolor{mygray}{STFormer + Graph-module} &{37.10, 0.980} &{32.96, 0.972} &{43.80, 0.990}  &{45.84, 0.995}  &{32.11, 0.958} &{31.86, 0.975} &37.28, 0.978 &1.02\\
    \bottomrule
  \end{tabular}}
  \end{center}
  \label{Table:sim-general}
\end{table*}

\begin{figure*}[thbp!]
\centering
\includegraphics[width=1.0\textwidth]{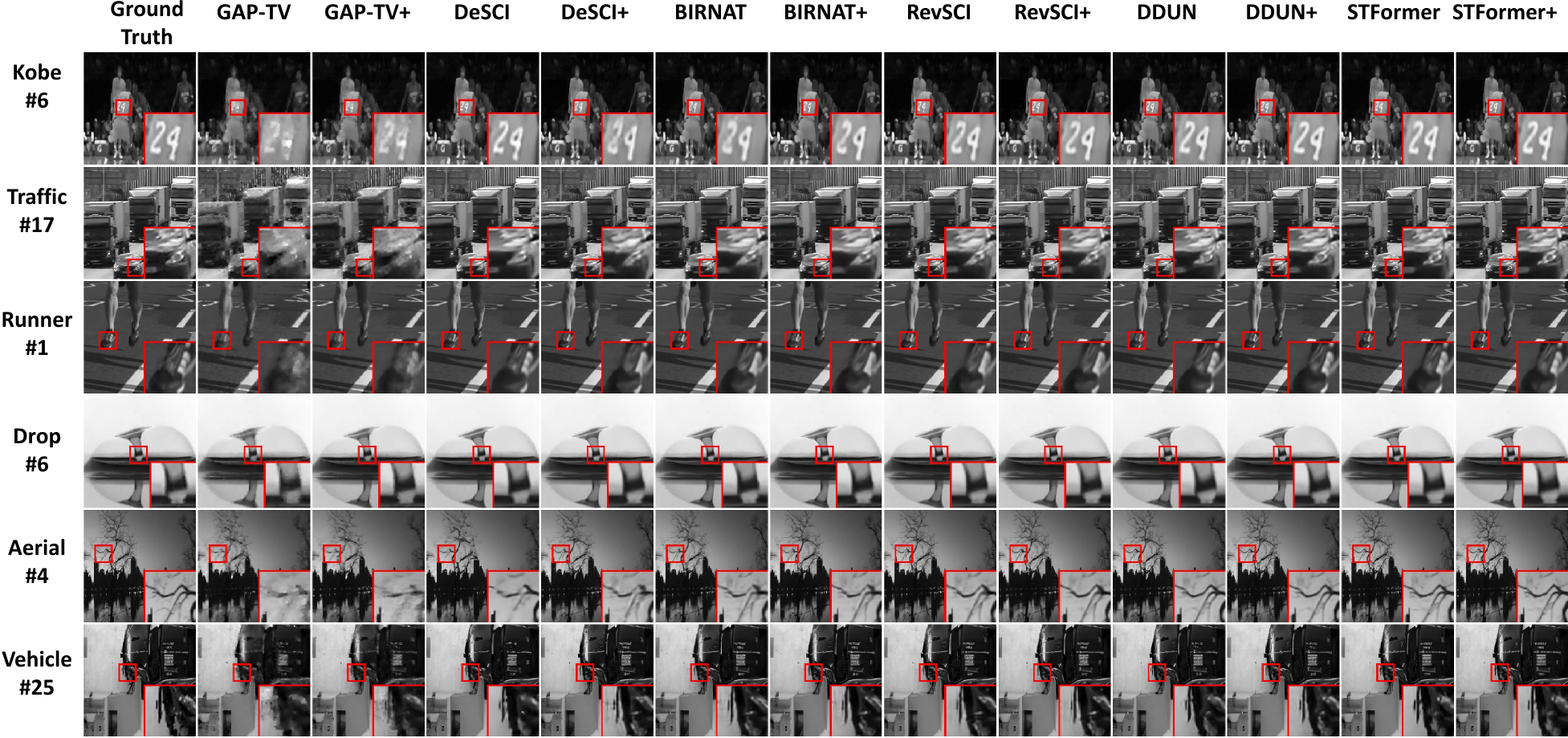}
\caption{\label{fig:sim-general}{The reconstructed frames of six benchmark grayscale simulation data. For simplicity, the `+' on the right side of each method indicates the results extended with our proposed graph module.}}
\end{figure*}

Revisiting Fig~\ref{fig:model}, the dynamic graph in our proposed MadyGraph aims to learn the complementary details of the initial prediction. This motivates us to explore the generalizability of our MadyGraph on various backbone models.
To valid the generalizability of MadyGraph, we impose our proposed dynamic graph module (the lower branch in Fig~\ref{fig:model}) on six popular backbone models: two iterative optimization-based methods GAP-TV~\cite{Yuan16ICIP_GAP} and DeSCI~\cite{Liu18TPAMI}, three deep networks BIRNAT~\cite{Cheng20ECCV_BIRNAT}, RevSCI~\cite{Cheng2021_CVPR_ReverSCI}, STFormer~\cite{wang2022spatial}, and an unfolding network DDUN~\cite{wu2021SCI3D}. Note that we directly append the well-trained dynamic graph module to these backbones to compute the residual details, and then add them together to obtain the final fine-grained videos without any fine-tuning.

The results on six widely used benchmark simulated grayscale data are given in Table~\ref{Table:sim-general} and Fig.~\ref{fig:sim-general}.
It can be observed that our proposed model consistently leads to performance boosting on the backbone models, \ie, providing \{2.71dB, 1.67dB, 0.52dB, 0.51dB, 0.92dB, 1.04dB\} higher PSNR than GAP-TV, DeSCI, BIRNAT, RevSCI, DDUN, and STFormer, respectively, 
with only a minimal increase in running time (approximately 0.1 s).
This verifies both the generalizability and efficiency of the proposed method for improving SCI reconstruction. {An interesting observation is that incorporating our graph module into the optimization-based methods, such as GAP-TV~\cite{Yuan16ICIP_GAP} and DeSCI~\cite{Liu18TPAMI}, could lead to significant performance-boosting (2.71dB and 1.67dB). For the DDUN~\cite{wu2021SCI3D} containing both the optimization and deep learning method, the performance improvement is 0.92 dB, even with a good baseline of 35.36 dB. We contribute this phenomenon to the complementary nature of our method, which possesses the strong representation capability of non-local pattern modeling. Our model could effectively leverage these advantages into optimization-based methods. Moreover, the proposed model achieves higher gains on worse initial predictions because the re-masking operation could significantly rectify the worse predictions, by extracting complementary information from the measurements and masks.}
Fig.\ref{fig:sim-general} plots selected reconstruction frames of different algorithms.
Compared with the counterparts, cleaner and sharper reconstruction corners and finer details with less noise are provided by appending our proposed graph model.

{Moreover, we list the results of changing the compression ratio $B$ in Figure~\ref{fig:multiB}. We set the compression ratio from 4 to 32. The flexible backbone methods, including GAP-TV, PnP-FFDNet, and PnP-FastDVDNet, are selected as the initial predictors due to their adaptability to multiple compression ratios without the need for additional training.  Concurrently, we retrained the MadyGraph across a range of compression ratios. Figure~\ref{fig:multiB} illustrates that the incorporation of the proposed graph module consistently enhances the performance of these backbone methods across various compression ratios. }

\begin{figure}[!t]
\centering
\includegraphics[width=0.9\columnwidth]{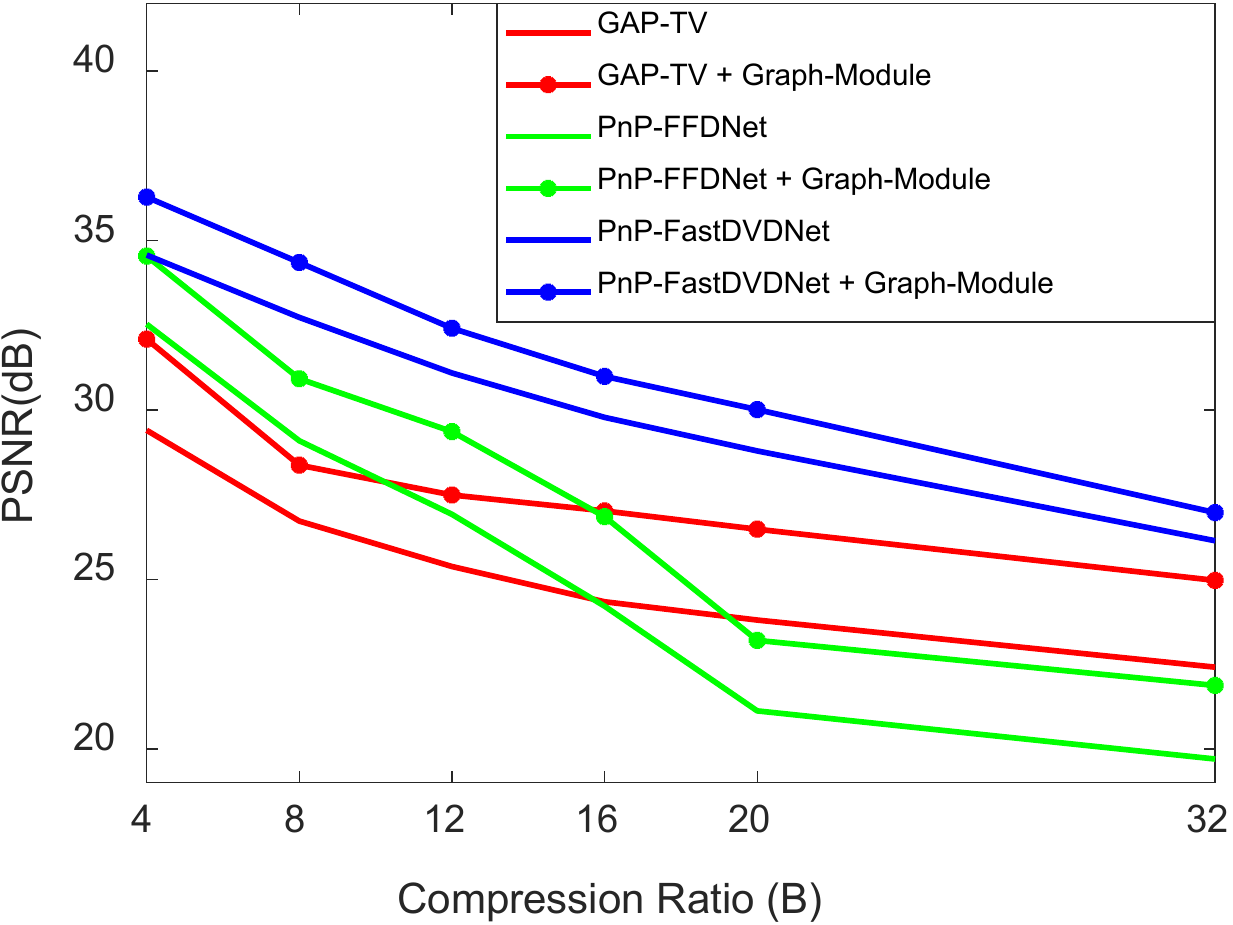}
\caption{\label{fig:multiB} {Reconstruction quality (average PSNR in dB of six benchmark data) varying compression rates B from 4 to 32 of the proposed Graph-Module based on different initial predictors.}}
\end{figure}

\begin{figure*}[t!]
\begin{center}
\includegraphics[width=0.9\textwidth]{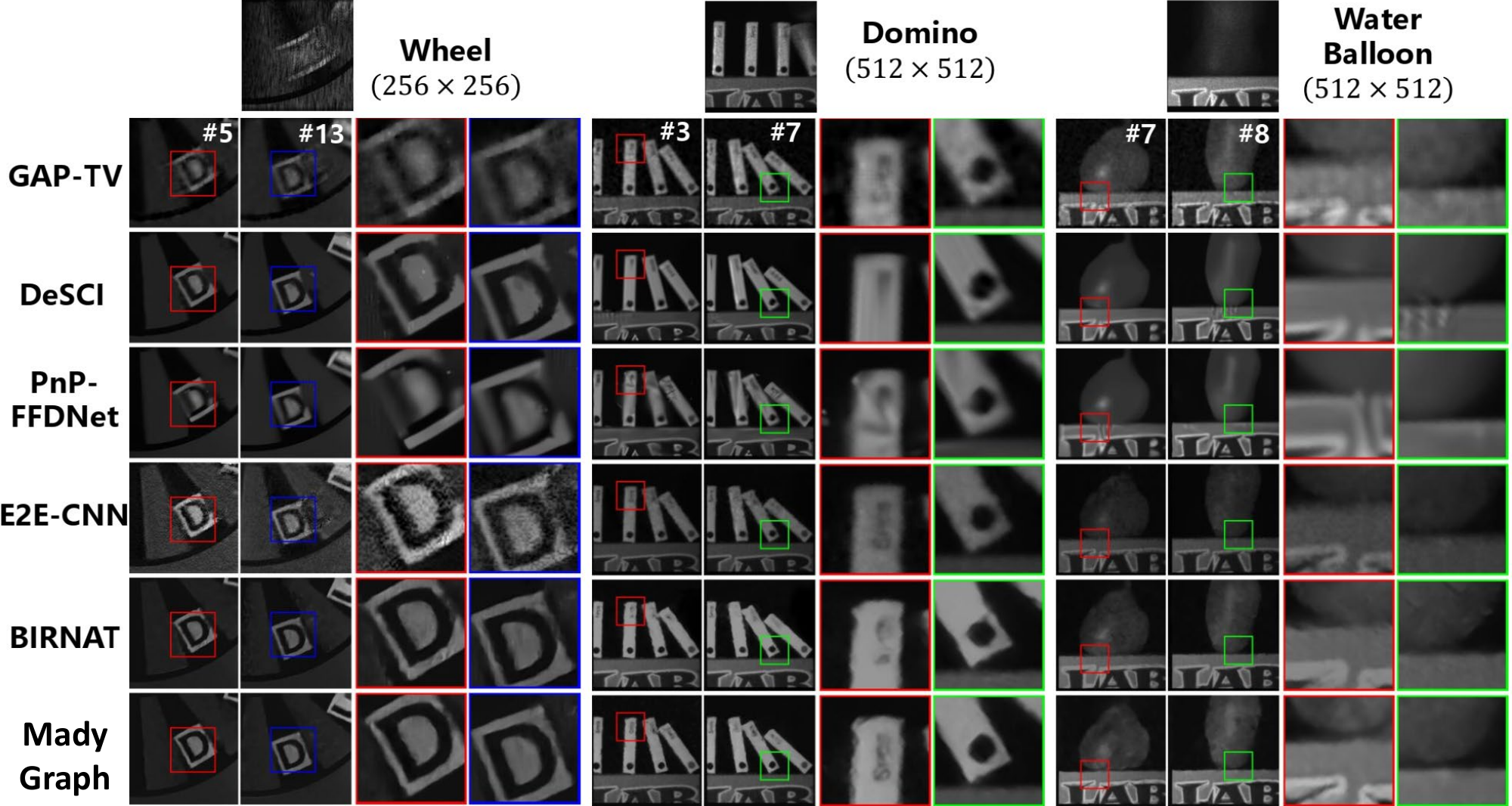}
\caption{\label{fig:real_data}{The reconstructed frames of real SCI data \texttt{Wheel}, \texttt{Domino}, and \texttt{Water Balloon}. }}
\end{center}
\end{figure*}

\begin{figure*}[t!]
\begin{center}
\includegraphics[width=0.9\textwidth]{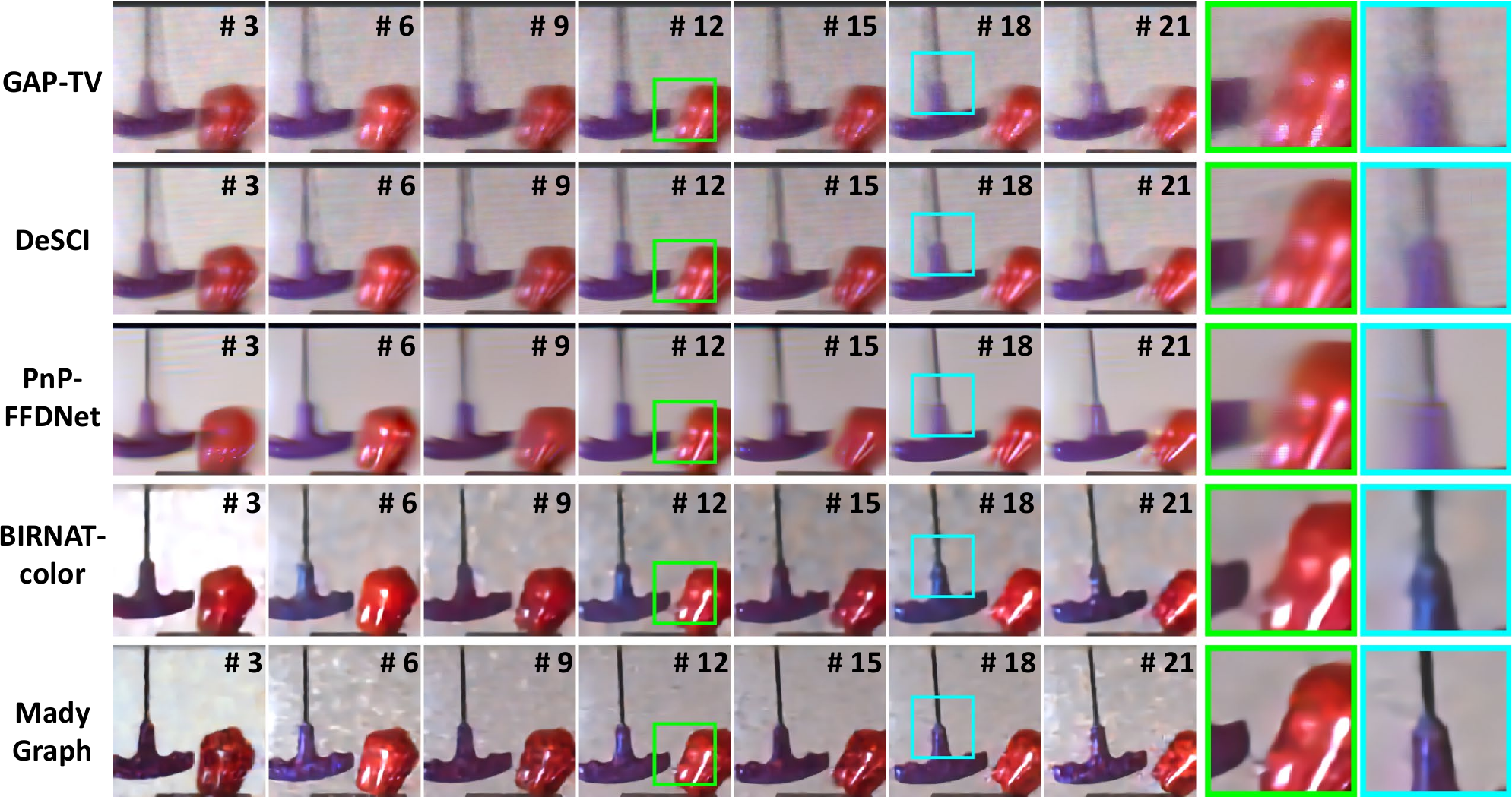}
\caption{\label{fig:color-real_data} {The reconstructed frames of real color SCI data \texttt{Hammer} with size $512 \times 512 \times 22$.}}
\end{center}
\end{figure*}

\subsection{Results on Real SCI Data}

To further demonstrate the robustness of our proposed MadyGraph, we apply it to real data captured by SCI cameras~\cite{Patrick13OE, Yuan14CVPR}. It’s worth noting that unlike simulated data which is considered noise-free, real data inevitably contains noise, making SCI reconstruction more challenging. To evaluate our model, we perform MadyGraph on three grayscale real data sets, \ie, \texttt{Wheel}~\cite{Patrick13OE}, \texttt{Domino}, and \texttt{Water Balloon}~\cite{Qiao2020_APLP}, as well as a color real data sets \texttt{Hammer}~\cite{Yuan14CVPR}. 
We retrain the networks of specific SCI systems using simulated measurements synthesized from real masks and video frames cropped from the DAVIS2017 dataset.

\subsubsection{Grayscale Real High-Speed Video}

We show the visualization results of reconstructed real videos in Fig.~\ref{fig:real_data}, compared with several representative reconstruction algorithms, namely
GAP-TV~\cite{Yuan16ICIP_GAP}, DeSCI~\cite{Liu18TPAMI}, PnP-FFDNet~\cite{Yuan20PnPSCI}, E2E-CNN~\cite{Qiao2020_APLP}, and BIRNAT~\cite{Cheng20ECCV_BIRNAT}. The key observations can be summarized as {follows}:
$i$) With the \texttt{Wheel} snapshot measurement of size 256$\times$256, we can recover 14 high-speed frames. As shown in Fig.~\ref{fig:real_data}, the generated letter `D' of MadyGraph shows clearer and smoother edges with fewer artifacts, while the comparison methods GAP-TV, E2E-CNN, and PnP-FFDNet result in some unpleasant artifacts. Compared with other methods, the boundary of reconstructed frames generated by MadyGraph is more consistent and smooth.
$ii$)For validation on the large-scale scenes, we recover the snapshot measurements \texttt{Domino} and \texttt{Water Balloon} as two videos of size 512$\times$512$\times$10. It can be observed that the proposed model generates sharper edges with less noise and provides more accurate contours, owing to the efficient modeling for non-local spatial-temporal dependencies. 
$iii$) {Meanwhile, the proposed method significantly saves testing time compared with optimization-based DeSCI (several hours), thus paving the way for real applications of SCI.}

\subsubsection{Color Real Video}

We evaluate the performance of our proposed MadyGraph model on a color real data \texttt{Hammer}, captured by a real SCI system. We convert the single Bayer mosaic measurement, which has a spatial resolution of 512$\times$512, into a video with 22 frames.
Our visualization results, depicted in Fig.~\ref{fig:color-real_data}, illustrate that MadyGraph outperforms several representative reconstruction algorithms - i.e., GAP-TV~\cite{Yuan16ICIP_GAP}, DeSCI~\cite{Liu18TPAMI}, PnP-FFDNet~\cite{Yuan20PnPSCI}, and BIRNAT~\cite{cheng2022recurrent}, in terms of restoring sharper edges when zooming in on local areas.
Specifically, GAP-TV and DeSCI exhibit some artifacts and blurred sketches in their reconstruction results, while our proposed MadyGraph is capable of restoring sharper edges and clearer profiles in comparison. In particular, the image obfuscation has significantly been reduced, proving the modeling power of the graph model for SCI.

In conclusion, the reconstructed videos of MadyGraph exhibit finer and more complete details compared to other methods, while maintaining an acceptable reconstruction time. These results suggest the applicability and efficiency of our algorithm in real-world settings.

\section{Conclusions}
In this paper, we present MadyGraph, a motion-aware dynamic graph, for snapshot compressive imaging. The proposed model aims to capture non-local meaningful related dependencies regardless of the distance in space and time, under the guidance of the motion information across video frames. Our proposed MadyGraph first provides initial prediction, then decodes the predicted video and physical masks together, successively performs the motion-aware dynamic node sampling to establish the graph model, and finally aggregates fine-grained residual information for video reconstruction. The experimental results on both simulation, real, grayscale, and color SCI data demonstrate the effectiveness and efficiency of the proposed MadyGraph. By exploring the non-local spatial-temporal correlations across video frames, MadyGraph achieves superior results on multiple video
SCI reconstruction tasks, particularly in complex and high-speed motion scenes. Furthermore, the graph module can be flexibly generalized to the other models, while enjoying fast inference speed.

However, there still remain several gaps between practical applications and current research~\cite{zhang2022compressive,suo2023computational}. One challenge lies in the fact that current reconstruction models mainly follow a `one-to-one' pipeline, where the well-trained models only match with a single SCI system.
A promising and challenging research direction is the ‘one-to-many’ setting, where a single model can effectively reconstruct videos from different SCI systems without additional training.
Another critical challenge is the increasingly severe computational burden posed by larger deep models, particularly in the context of real-time video reconstruction on edge devices or small terminal devices.
With powerful modeling capabilities, MadyGraph may potentially drive the solution to those challenges in future research~\cite{Wang_2023_CVPR}.

{Overall, we hope to introduce a novel way for modeling non-local spatial-temporal correlations for video {compressive imaging}. Furthermore, MadyGraph can potentially extend to a wider range of application scenarios in the future, wherever non-local modeling plays a pivotal role.}

\ifCLASSOPTIONcompsoc
  \section*{Acknowledgments}
\else
  \section*{Acknowledgment}
\fi

Xin Yuan would like to thank the National Natural Science Foundation of China [62271414], Zhejiang Provincial Natural Science Foundation of China [LR23F010001], and the Research Center for Industries of the Future (RCIF) at Westlake University for supporting this work.
Bo Chen would like to thank the National Natural Science Foundation of China [U21B2006], the Shaanxi Youth Innovation Team Project, the Fundamental Research Funds for the Central Universities [QTZX24003 and QTZX22160]; and the 111 Project [B18039].
Ruiying Lu would like to thank the Natural Science Basic Research Plan in Shaanxi Province of China under Grant [2024JC-YBQN-0661].

{\small
\bibliographystyle{IEEEtran}
\bibliography{Madygraph}
}

\end{document}